\documentclass{article}

\usepackage[final]{neurips_2025}

\usepackage[utf8]{inputenc} % allow utf-8 input
\usepackage[T1]{fontenc}    % use 8-bit T1 fonts
\usepackage{hyperref}       % hyperlinks
\usepackage{url}            % simple URL typesetting
\usepackage{booktabs}       % professional-quality tables
\usepackage{amsfonts}       % blackboard math symbols
\usepackage{nicefrac}       % compact symbols for 1/2, etc.
\usepackage{microtype}      % microtypography
\usepackage{xcolor}         % colors

\newif\ifcomments
\commentstrue
\ifcomments
    \newcommand{\todo}[1]{{\textcolor[rgb]{1.0,0.0,0.0}{[TODO: {\it #1}]}}}
    \newcommand{\caelan}[1]{\textcolor{blue}{(CG: #1)}}
    \newcommand{\ajay}[1]{\textcolor{red}{(AM: #1)}}
    \newcommand{\shuo}[1]{\textcolor{green}{(SC: #1)}}
    \newcommand{\danfei}[1]{\textcolor{cyan}{(DX: #1)}}
    \newcommand{\liqian}[1]{\textcolor[rgb]{0.5,0.0,1.0}{(LM: #1)}}
\else
    \newcommand{\todo}[1]{}
    \newcommand{\caelan}[1]{}
    \newcommand{\ajay}[1]{}
    \newcommand{\shuo}[1]{}
    \newcommand{\danfei}[1]{}
    \newcommand{\liqian}[1]{}
\fi

\usepackage{graphicx} 
\usepackage{amsmath, amssymb}
\usepackage{subcaption}
\usepackage{multirow}

\usepackage{algorithm}
\usepackage{algpseudocode}
\usepackage{wrapfig}
\usepackage{booktabs}

\usepackage{adjustbox}

\title{
Generalizable Domain Adaptation for \\
Sim-and-Real Policy Co-Training
}

\author{
\begin{tabular}{ccc}
Shuo Cheng\textsuperscript{1*} & Liqian Ma\textsuperscript{1*} & Zhenyang Chen\textsuperscript{1} \\
Ajay Mandlekar\textsuperscript{2\textdagger} & Caelan Garrett\textsuperscript{2\textdagger} & Danfei Xu\textsuperscript{1}
\end{tabular} \\[6pt]
\textsuperscript{1} Georgia Institute of Technology \quad
\textsuperscript{2} NVIDIA Corporation \\[4pt]
\textsuperscript{*} and \textsuperscript{\textdagger} denote equal contribution \\[4pt]
\texttt{\{shuocheng, mlq\}@gatech.edu}
}

\begin{document}

\maketitle

\begin{abstract}
Behavior cloning has shown promise for robot manipulation, but real-world demonstrations are costly to acquire at scale. While simulated data offers a scalable alternative, particularly with advances in automated demonstration generation, transferring policies to the real world is hampered by various simulation and real domain gaps. In this work, we propose a unified sim-and-real co-training framework for learning generalizable manipulation policies that primarily leverages simulation and only requires a few real-world demonstrations. Central to our approach is learning a domain-invariant, task-relevant feature space. Our key insight is that aligning the \emph{joint distributions} of observations and their corresponding actions across domains provides a richer signal than aligning observations (marginals) alone. We achieve this by embedding an Optimal Transport (OT)-inspired loss within the co-training framework, and extend this to an Unbalanced OT framework to handle the imbalance between abundant simulation data and limited real-world examples. We validate our method on challenging manipulation tasks, showing it can leverage abundant simulation data to 
achieve up to a 30\% improvement in the real-world success rate and even generalize to scenarios seen \emph{only in simulation}. Project webpage: \url{https://ot-sim2real.github.io/}.

\end{abstract}

\section{Introduction}

% \begin{itemize}
%     \item \url{https://gtvault-my.sharepoint.com/:p:/g/personal/zchen927_gatech_edu/EQMWZBOvfbNOjWhzHkDRtigBz5DNynWZEWG72awGxC_Uwg?e=8buoCn}
% \end{itemize}

Behavior cloning~\cite{pomerleau1988alvinn} is a promising approach for acquiring robot manipulation skills directly in the real world, due to its simplicity and effectiveness in mimicking expert demonstrations~\cite{robomimic2021, florence2022implicit}. However, achieving robust and generalizable performance requires collecting large-scale datasets~\cite{khazatsky2024droid, open_x_embodiment_rt_x_2023} across diverse environments, object configurations, and tasks. This data collection process is labor-intensive, time-consuming, and costly, posing significant challenges to scalability in real-world applications.

Recently, with rapid advancements in physics simulators~\cite{Genesis, Xiang_2020_SAPIEN}, procedural scene generation~\cite{raistrick2024infinigen, deitke2022️}, and motion synthesis techniques~\cite{mandlekar2023mimicgen, cheng2023nod}, there has been growing interest in leveraging simulation as an alternative source of training data. These simulation-based approaches enable scalable and controllable data generation, allowing for diverse and abundant supervision at a fraction of the real-world cost. However, transferring policies trained in simulation to the physical world remains a non-trivial challenge due to sim-to-real gap—the discrepancies between the simulated and real-world environments that a policy encounters during execution. These differences can manifest in various forms, such as variations in visual appearance, sensor noise, and action dynamics~\cite{andrychowicz2020learning, tobin2017domain}. In particular, learning visuomotor control policies that remain robust under changing perceptual conditions during real-world deployment continues to be an open area of research.

\begin{figure}[h]
    \centering
    \includegraphics[width=1.0\linewidth]{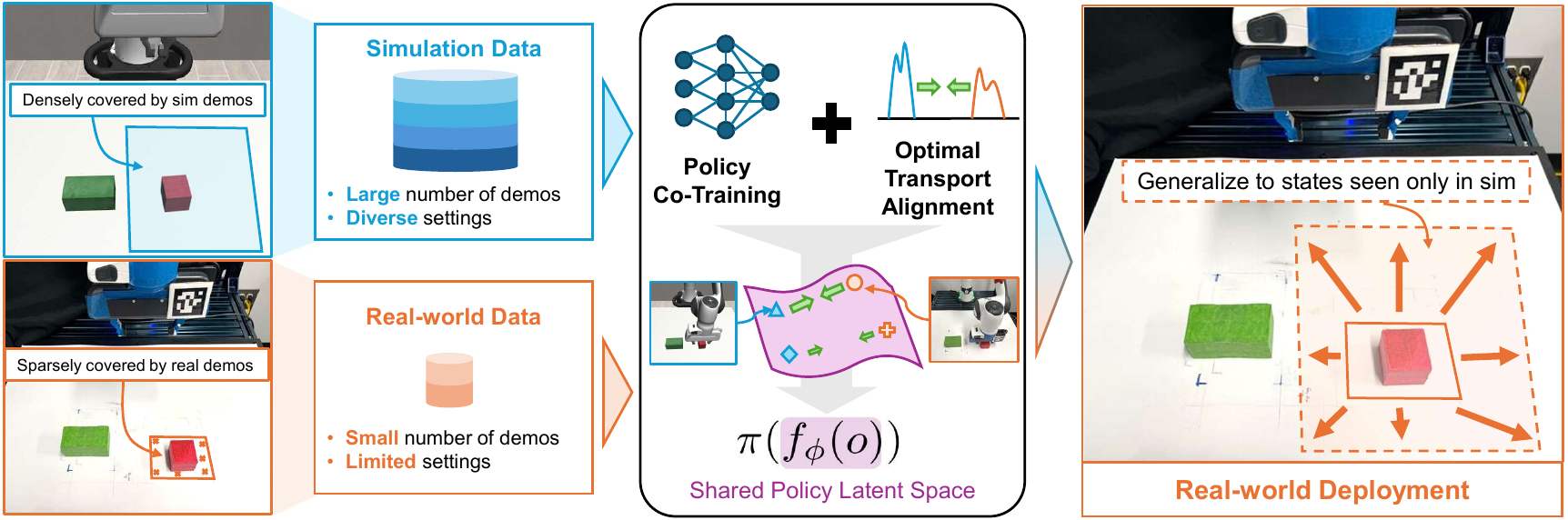}
    \caption{\textbf{Sim-and-Real Co-Training with Optimal Transport}. We use behavior cloning to train a real-world policy from sparse real-world and dense simulation demos. Leveraging Optimal Transport to align feature spaces, our method enables generalization to scenarios seen only in simulation. }
    \label{fig:teaser}
    % \vspace{-3pt}
\end{figure}

Common strategies to bridge this domain gap include domain randomization~\cite{andrychowicz2020learning, tobin2017domain} and data augmentation~\cite{hansen2020self, yarats2021mastering}, though these often require careful tuning. Domain adaptation (DA) techniques aim to explicitly align distributions, either at pixel~\cite{bousmalis2017unsupervised, james2019sim} or feature levels~\cite{tzeng2014deep, long2015learning, zhao2019learning}. However, many feature-level methods align only marginal observation distributions (e.g., MMD~\cite{tzeng2014deep, long2015learning}), which can be insufficient for fine-grained manipulation alignment as it may not preserve action-relevant relationships across domains. More recently, sim-and-real co-training—simply training a single policy on mixed data from both domains~\cite{wei2025empirical, maddukuri2025sim}—has shown surprising effectiveness. We argue that while beneficial for data diversity, such co-training approaches typically lack explicit constraints for feature space alignment across domains, potentially hindering optimal transfer and generalization because they don't enforce a consistent mapping of task-relevant structures.

We present a unified sim-and-real co-training framework that explicitly learns a shared latent space where observations from simulation and the real world are aligned and preserve action-relevant information. Our key insight is that aligning the \emph{joint distributions} of observations and their corresponding actions or task-relevant states across domains provides a direct signal for learning transferable features. 
%than aligning observation distributions alone. 
Concretely, we leverage Optimal Transport (OT)~\cite{courty2016optimal} as an alignment objective to learn representations where the geometric relationships crucial for action prediction are consistent, irrespective of whether the input comes from simulation or the real world. Further more, to robustly handle the \emph{data imbalance} in co-training with abundant simulation data and limited real-world data, we further extend to an Unbalanced OT (UOT) formulation~\cite{fatras2021unbalanced, chizat2018scaling} and develop a temporally-aware sampling strategy to improve domain alignment learning in a mini-batch OT setting. 

\textbf{Our contributions are}: (1) a sim-and-real co-training framework that learns a domain-invariant yet task-salient latent space to improve real-world performance with abundant simulation data, (2) an Unbalanced Optimal Transport framework and temporally-aware sampling strategy to mitigate data imbalance and improve alignment quality in mini-batch OT training, (3) comprehensive experiments using both image and point-cloud modalities, evaluating sim-to-sim and sim-to-real transfer across diverse manipulation tasks, demonstrating up to a 30\% average success rate improvement and achieving generalization to real-world scenarios for which the training data only appears in simulation.

\section{Related Work}
\textbf{Behavior Cloning for Robot Manipulation.}
Behavior cloning (BC) trains policies to map observations to actions by imitating expert demonstrations~\cite{robomimic2021, zhao2023learning, chi2023diffusion, Ze2024DP3}, offering an effective path to human-like manipulation skills. Generalization heavily depends on dataset diversity. While some efforts focus on large-scale real-world data collection~\cite{open_x_embodiment_rt_x_2023, khazatsky2024droid} or more efficient collection techniques~\cite{wu2024gello, chi2024universal, kareer2024egomimic}, this remains costly and time-consuming. Physical simulators provide a low-cost alternative, with automatic motion synthesis leveraging privileged information to generate large-scale simulated demonstrations~\cite{mandlekar2023mimicgen, jiang2024dexmimicgen, cheng2024nodtamp}. Our work combines abundant simulated data with few real-world demonstrations to train robust BC policies.

% Behavior cloning is an effective supervised learning approach for robots to acquire human-like manipulation skills, in which a policy is trained to directly map observations to actions by imitating expert demonstration data~\cite{robomimic2021, zhao2023learning, chi2023diffusion, Ze2024DP3}. The generalization capability of behavior cloning policies largely depends on the diversity of the training dataset. To address this, recent efforts have focused on acquiring diverse datasets directly in the real world either through large-scale, distributed data collection~\cite{open_x_embodiment_rt_x_2023, khazatsky2024droid}, or by developing more efficient data collection techniques~\cite{wu2024gello, chi2024universal, kareer2024egomimic}. However, real-world data collection is both costly and time-consuming, as it typically requires access to real robots and human teleoperators. In contrast, physical simulators offer a low-cost, safe alternative for data collection. By leveraging privileged information available only in simulation, automatic motion synthesis techniques can efficiently generate large-scale simulated demonstrations~\cite{mandlekar2023mimicgen, jiang2024dexmimicgen, cheng2024nodtamp}. In this work, we leverage a large-scale simulation dataset in combination with a small number (e.g., 10) of real-world demonstrations to train a behavior-cloning policy that can be robustly deployed in the real world.

\textbf{Sim-to-real Transfer and Co-training.} 
% \caelan{We sometimes use a hyphen in cotraining and other times don't}
Policies trained solely in simulation often underperform in the real world due to the sim-to-real gap—discrepancies in visual appearance and dynamics. For quasi-static manipulation, the visual domain gap is typically the primary bottleneck.  
Domain randomization exposes policies to varied simulated visual conditions to build robustness to real-world variability~\cite{tobin2017domain, james2017transferring, yuan2024learning}. However, its success depends on how well randomized parameters cover true real-world distributions, often requiring manual tuning.
Domain adaptation (DA) explicitly aligns source (simulation) and target (real) domains~\cite{farahani2021brief,dan2025x}. Pixel-level DA uses image translation to make simulated images resemble real ones~\cite{bousmalis2017unsupervised, james2019sim, ho2021retinagan}. Feature-level DA, which is often more scalable for end-to-end learning, focuses on learning domain-invariant representations~\cite{long2015learning,zhao2019learning,tzeng2014deep,raychaudhuri2021cross,kim2020domain}.
Sim-and-real co-training, where a policy is jointly trained on mixed data~\cite{wei2025empirical, maddukuri2025sim}, offers a simple and effective alternative. While co-training enhances generalization through data diversity, it typically lacks explicit constraints to align learned feature spaces across domains. We build upon co-training by incorporating feature-level domain adaptation via Optimal Transport to promote latent space alignment, thereby improving real-world policy performance.

\textbf{Optimal Transport for Domain Adaptation.} Optimal Transport (OT) offers a principled framework for aligning distributions, widely adopted for domain adaptation~\cite{courty2016optimal, courty2014domain, perrot2016mapping, redko2017theoretical,courty2017joint,nguyen2024dude,kedia2025one}. Traditional OT methods compute a transport plan between source and target samples, then train a new model on the transported source. Most relevant to us is DeepJDOT~\cite{damodaran2018deepjdot}, which builds on JDOT~\cite{courty2017joint} to align joint distributions of features and (pseudo) labels for unsupervised domain adaptation, where target labels are unavailable. Our work builds on these principles for sim-and-real co-training imitation. Unlike unsupervised DA, we leverage available action or state labels from limited real-world demonstrations as ``soft'' supervision to guide a more task-relevant alignment of joint observation-label distributions across domains. To robustly handle the inherent data imbalance between abundant simulation and scarce real data, we incorporate an Unbalanced OT (UOT) loss~\cite{fatras2021unbalanced} into our co-training framework and develop a temporally-aware sampling strategy to improve mini-batch UOT training.

% \caelan{Looks good}

% \textbf{Optimal Transport for Domain Adaptation.} Optimal Transport (OT), a general framework for measuring distribution distances, has been widely adopted for domain adaptation~\citep{courty2016optimal, courty2014domain, perrot2016mapping,redko2017theoretical}. These methods focus on computing a coupling—or transport plan—that maps samples from the source distribution to the target domain. Once the alignment is established, a new classifier or regressor is trained on the transported source data. Applied to Deep Learning, DeepJDOT~\cite{damodaran2018deepjdot} relaxes the exact OT assumption and use the differentiable Sinkhorn distance~\cite{} to learn feature representations that enables domain adaptation. Building on these insights, we incorporate an Unbalanced OT~\cite{fatras2021unbalanced} loss into behavior cloning framework and propose a temporally-aware sampling strategy to further mitigate data imbalance in sequential decision-making tasks. This enhances the robustness and effectiveness of domain alignment in learning from demonstrations.

\vspace{-10pt}

\section{Preliminaries and Problem Setting}
\label{sec:prelim}
\vspace{-10pt}
%\subsection{Optimal Transport} \label{ssec:ot}
% \danfei{I'd still keep it here. This is what I'd expect from a robot learning paper -- describe the prerequisite}

Our method builds on the principle of Optimal Transport (OT) for aligning two empirical distributions. Let $\mathbf{U}=\{u_i\}_{i=1}^{n}$ and $\mathbf{V} = \{v_j\}_{j=1}^{m}$ represent the data points drawn from a {\em source} domain and a {\em target} domain, with corresponding empirical distributions $\mathbf{p}=\sum_{i=1}^{n}p_i\delta_{u_i}$ and $\mathbf{q}=\sum_{j=1}^{m}q_j\delta_{v_j}$. We define the ground cost matrix ${C}=(C_{i,j}) \in \mathbb{R}^{n\times m}$ with $C_{i, j}=c(u_i, v_j)$, where $c(\cdot,\cdot)$ is a cost function, which is often defined as squared Euclidean distance. Optimal Transport (OT) seeks to find an optimal plan $\Pi$ that maps the distribution $\mathbf{p}$ to $\mathbf{q}$ that minimizes the displacement cost $W_c(\mathbf{p}, \mathbf{q})$:
% \begin{equation}
% {W}_{c}(\mathbf{p}, \mathbf{q}) = \min_{\Pi \in \Gamma(\mathbf{p}, \mathbf{q})} \langle\Pi, C\rangle_{F}, \; \text{s.t.} \; \Gamma(\mathbf{p}, \mathbf{q}) = \{\Pi \in \mathbb{R}_{+}^{n\times m}\mid \Pi\mathbf{1}_m = \mathbf{p}, \Pi^{\top}\mathbf{1}_n=\mathbf{q}\}.
% \label{eq:ot}
% \end{equation}
% \caelan{What about this simplified version instead?}
\begin{equation}
{W}_{c}(\mathbf{p}, \mathbf{q}) = \min_{\Pi \in \mathbb{R}_{+}^{n\times m}} \langle\Pi, C\rangle_{F}, \; \text{s.t.} \; \Pi\mathbf{1}_m = \mathbf{p}, \Pi^{\top}\mathbf{1}_n=\mathbf{q}.
\label{eq:ot}
\end{equation}
%Here $\Gamma(\mathbf{p}, \mathbf{q})$ denotes distribution marginals $\mathbf{p}$ and $\mathbf{q}$, and $\langle\cdot,\cdot\rangle_{F}$ is the Frobenius dot product. 

\vspace{-5pt}
\subsection{Problem Setting: Sim-and-Real Policy Co-Training}
\label{subsec:problem_setting}
We address the challenge of learning robust real-world robotic manipulation policies $\pi$. Our approach minimizes the need for extensive real-world data collection by primarily leveraging abundant simulation data alongside a small set of real-world demonstrations. This is framed as a sim-and-real co-training problem \cite{maddukuri2025sim,matas2018sim}, where a single policy is trained on data from both domains. Specifically, we consider a source domain (simulation, denoted $src$) and a target domain (real-world, denoted $tgt$). 
We model the domains as Partially Observable Markov Decision Processes (POMDPs) that share an underlying, generally unobserved, state space $\mathcal{S}$ and an action space $\mathcal{A}$. The policy receives observations comprising high-dimensional visual input $o \in \mathcal{O}$ (e.g., RGB images, 3D point clouds) generated by the emission function $E: \mathcal{S} \mapsto \mathcal{O}$, together with low-dimensional proprioceptive information $x \in \mathcal{X}$ (e.g., robot joint angles, end-effector pose).

\textbf{Domain Gaps.} The central challenge is the domain gap, particularly the \textit{visual observation gap}. For the same underlying robot and environment state $s \in \mathcal{S}$, visual observations emitted in simulation, $o_{src} = E_{src}(s)$, can differ significantly from those in the real world, $o_{tgt} = E_{tgt}(s)$. This discrepancy arises from factors like variations in visual appearance (textures, lighting), sensor noise, and rendering artifacts (e.g., differences between simulated ray casting and real-world light transport). 
As a result, the marginal observation distributions differ between the domains, namely $P_{src}(o_{src}) \neq P_{tgt}(o_{tgt})$.
In contrast, actions $a$ and proprioceptive states $x$ are assumed to be largely consistent for a given $s$ due to consistent data generation strategies (discussed next) and accurate robot state estimation. While differences in dynamics also contribute to the domain gaps, our focus on learning quasi-static prehensile manipulation tasks from human-sourced demonstrations means that the dynamics gap is typically less dominant than the observation gap. 
%We detail how we address the residual dynamics gap in Appendix~\ref{TBD}.

\textbf{Data Sources and Data Imbalance.}
In the source (simulation) domain, we leverage the ability to automatically generate a large dataset of $N_{src}$ trajectories, $D_{src} = \{(o^i_{src}, x^i_{src}, a^i_{src})\}_{i=1}^{N_{src}}$. Specifically, we leverage an automated demonstration generation tool MimicGen \cite{mandlekar2023mimicgen}, which utilizes privileged information available in simulation to create a diverse set of experiences covering a broad underlying state space, $S_{src}$. MimicGen populates synthetic demonstrations based on a handful of human demonstrations, which ensures that the generated data is behaviorally consistent with human demonstrations.
%, an assumption important to our domain alignment framework. 
%Data collection in the target (real-world) domain is resource-intensive. 
We collect a limited number of $N_{tgt}$ demonstrations, $D_{tgt} = \{(o^j_{tgt}, x^j_{tgt}, a^j_{tgt})\}_{j=1}^{N_{tgt}}$ (where $N_{src} \gg N_{tgt}$), typically through human teleoperation. These real-world demonstrations will naturally cover a much smaller and potentially distinct subset of states, $S_{tgt}$. This difference in data coverage leads to the challenge of \textbf{partial data overlap}. While we assume there is a region of states common to both $S_{src}$ and $S_{tgt}$ where direct alignment is possible, a significant portion of $S_{src}$ (our rich simulated data) will not have corresponding real-world demonstrations in $S_{tgt}$. Conversely, $S_{tgt}$ might contain details specific to real world (e.g., demonstration behaviors). Effectively leveraging the entirety of $D_{src}$ for real-world performance, especially for states outside the direct sim-real overlap in demonstrations, is the problem we address.

\textbf{Objective: Generalizable Domain Adaptation.} Our goal is to learn a single, generalizable policy $\pi_\theta(a | z, x)$ and an observation encoder $f_\phi: \mathcal{O}_{src} \cup \mathcal{O}_{tgt} \rightarrow \mathcal{Z}$. This encoder maps high-dimensional visual observations $o$ from both source and target domains to a shared latent space $\mathcal{Z}$. The primary objective is to achieve high policy performance in the target (real-world) domain, especially in scenarios not explicitly covered by the limited real-world demonstrations $D_{tgt}$. This entails \textbf{two objectives}. First, for states within the overlapping regions of $S_{src}$ and $S_{tgt}$, we aim to learn high-quality embeddings $z=f_\phi(o)$ that are well-aligned across domains, such that $f_\phi(o_{src}) \approx f_\phi(o_{tgt})$ in corresponding states, facilitating effective policy learning. This is also the assumption of most co-training methods~\cite{wei2025empirical,maddukuri2025sim}. Second, for states covered in $S_{src}$ but not in $S_{tgt}$, the encoder $f_\phi$ must produce embeddings $f_\phi(o_{tgt})$ for novel target observations that are consistent with the embeddings of their simulated counterparts $f_\phi(o_{src})$. This requires the learned representations to capture domain-invariant, task-relevant features, enabling policy to \emph{generalize to target (real-world) scenarios for which training data is only present in the source (simulated) domain}.

\section{Method}

\begin{figure}
    \centering
    \includegraphics[width=1\linewidth]{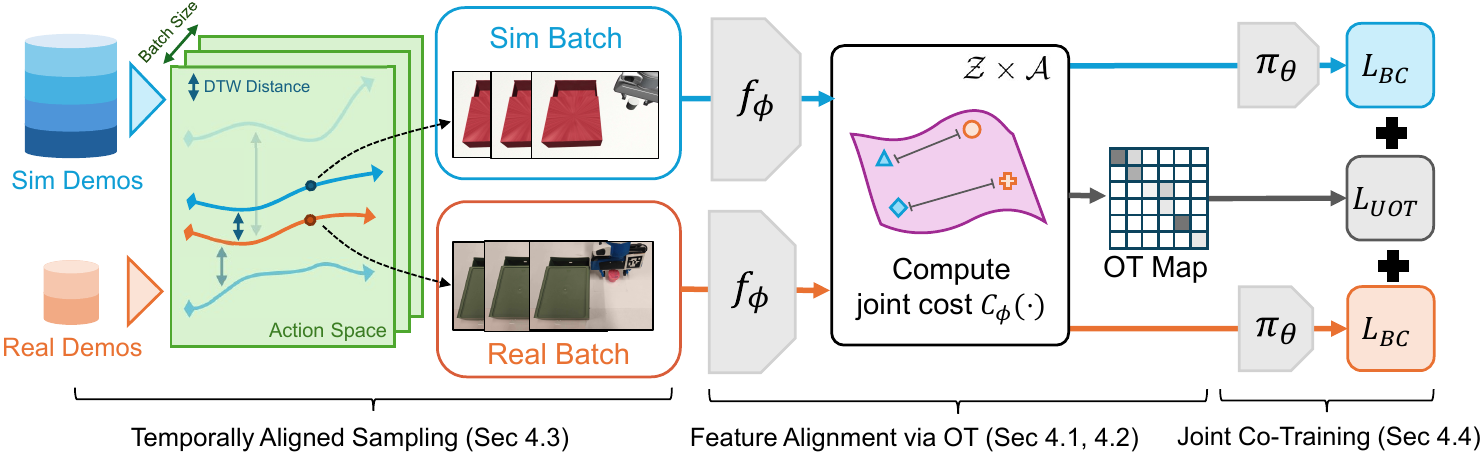}
    \caption{\textbf{Method Overview.} 
    % In order to account for size discrepancy between the simulation and real demos, we first use Dynamic Time Warping (DTW) to sample pairs of simulation and real demonstrations that are similar. Then we apply Unbalanced, to allow for imperfect matches, Optimal Transport (OT), where the matching costs incorporate in robot actions. Finally, we combine the OT loss with traditional Behavior Cloning (BS) losses for policy learning.
   Our sim-and-real co-training framework learns a domain-invariant
   %, task-relevant 
   latent space to improve real-world performance using abundant simulation demos and a small number of real-world demos. It leverages an Unbalanced Optimal Transport loss and a temporal sampling strategy to address data imbalance and improve alignment quality during mini-batch training.} 
    \label{fig:enter-label}
    \vspace{0pt}
\end{figure}

To learn a generalizable policy $\pi_\theta$ with robust feature $f_\phi$ from imbalanced sim-and-real datasets ($|D_{src}| \gg |D_{tgt}|$), as outlined in Section~\ref{sec:prelim}, we propose a co-training framework that explicitly aligns the latent representation through Optimal Transport (OT). Our core strategy is to leverage action information to guide the alignment of learned visual latent features ($z = f_\phi(o)$).  %  \in \mathcal{Z}
This alignment objective simultaneously encourages the encoder $f_\phi$ to discover domain-invariant representations while preserving detailed information for action prediction. Specifically, we propose a formulation based on OT to align joint observation-action distributions (Sec.~\ref{ssec:jot}). We further address the data imbalance problem through Unbalanced Optimal Transport (UOT) (Sec.~\ref{ssec:uot}) and a temporally-aware sampling strategy (Sec.~\ref{ssec:sampling}), all integrated into a unified co-training framework (Sec.~\ref{ssec:joint}).

\subsection{Optimal Transport for Action-Aware Feature Alignment}
\label{ssec:jot}

While standard co-training methods \cite{wei2025empirical, maddukuri2025sim} offer implicit feature alignment, and marginal distribution matching (e.g., Maximum Mean Discrepancy \cite{tzeng2014deep}) can overlook fine-grained correspondences, we seek a more structured approach. To learn domain-invariant visual features $z=f_\phi(o)$ that are also predictive of actions $a \in \mathcal{A}$, we propose to align the joint distributions $P_{src}(f_\phi(o_{src}), a_{src})$ and $P_{tgt}(f_\phi(o_{tgt}), a_{tgt})$ using Optimal Transport. This encourages the encoder $f_\phi$ to learn representations where the geometric relationships between (visual feature, action) pairs are preserved across domains. By minimizing an OT-based loss, the encoder $f_\phi$ is trained to shape the embedding space $\mathcal{Z}$ such that structures relevant to action prediction are consistent between simulation and the real world.

%\caelan{Previously $x$ was explicitly in the samples}
Formally, given source samples $\{(o^i_{src}, a^i_{src})\}_{i=1}^{N_{src}}$ and target samples $\{(o^j_{tgt}, a^j_{tgt})\}_{j=1}^{N_{tgt}}$, we aim to find an optimal transport plan $\Pi^*$ and an optimal encoder $f_\phi^*$ that minimize the transportation cost between their joint distributions in the $(z,a)$ space. Based on the general OT formulation in Eq.~\ref{eq:ot}, learning objective for the encoder $f_\phi$, and implicitly the transport plan $\Pi$, can be expressed as finding $f_\phi$ that minimizes the Wasserstein distance between $P_{src}(f_\phi(o_{src}), a_{src})$ and $P_{tgt}(f_\phi(o_{tgt}), a_{tgt})$:
%\begin{equation}
$\min_{f_\phi} W_C\left(P_{src}(f_\phi(o_{src}), a_{src}), P_{tgt}(f_\phi(o_{tgt}), a_{tgt})\right)$.
%\label{eq:ot_loss_general}
%\end{equation}
The ground cost $c$ ideally combine distances in both the learned visual latent space $\mathcal{Z}$ and the action space $\mathcal{A}$, for instance:
\begin{equation}
    C_{\phi}\left( (f_\phi(o^i_{src}), a^i_{src}), (f_\phi(o^j_{tgt}), a^j_{tgt}) \right) = \alpha_1 \cdot d_{\mathcal{Z}}(f_\phi(o^i_{src}), f_\phi(o^j_{tgt})) + \alpha_2 \cdot d_{\mathcal{A}}(a^i_{src}, a^j_{tgt}).
    \label{eq:za_ground_cost}
\end{equation}
Minimizing 
%Eq.~\ref{eq:ot_loss_general} 
this objective using an iterative algorithm like Sinkhorn~\cite{cuturi2013sinkhorn} creates a bi-level optimization. In the inner loop, for a fixed $f_\phi$, an approximately optimal transport plan $\Pi$ is computed. In the outer loop, $f_\phi$ is updated to reduce the cost incurred by this plan. This process effectively trains the encoder $f_\phi$ to produce embeddings $z$ that make the source and target joint distributions $(z,a)$ less costly to align. A key advantage of OT is its ability to preserve geometric structures; by guiding alignment with action similarity (via $d_{\mathcal{A}}$), we shape the embedding function $f_\phi$ to cluster visual observations that lead to similar actions, irrespective of their domain of origin.

\textbf{Practical Implementation: Proprioception as Guidance.} While direct alignment of $(z,a)$ is principled, discrepancies in controller characteristics or action representations between simulation and real-world teleoperation can make $d_{\mathcal{A}}(a_{src}, a_{tgt})$ an unreliable indicator of behavioral similarity. As a robust practical compromise, we leverage proprioceptive information $x \in \mathcal{X}$ (e.g., end-effector pose), which is more consistently represented across domains (Section~\ref{subsec:problem_setting}) and highly correlated with robot behavior. Thus, our implemented ground cost ${c}$ replaces actions $a$ with proprioceptive states $x$, which is used in our UOT formulation (detailed in Section~\ref{ssec:uot}).

\subsection{Unbalanced Optimal Transport for Robust Alignment}
\label{ssec:uot}

The standard OT formulation (Eq.~\ref{eq:ot}) enforces strict marginal constraints, requiring all mass from the source distribution to be transported to the target and vice-versa. This is problematic in our sim-to-real setting primarily due to significant data imbalance ($|D_{src}| \gg |D_{tgt}|$) and the partial overlap between the state spaces covered by $D_{src}$ and $D_{tgt}$ (Section~\ref{subsec:problem_setting}). Standard OT would either distort the latent space by forcing many-to-few mappings or create spurious alignments between non-corresponding states. To address these challenges, we employ Unbalanced Optimal Transport (UOT) \cite{chizat2018scaling, fatras2021unbalanced}. UOT relaxes the hard marginal constraints of OT by introducing regularization terms that penalize deviations, thereby allowing for partial mass transport. This enables UOT to selectively align subsets of the distributions that are most similar according to the ground cost, while effectively down-weighting or ignoring the transport for dissimilar or unmatched portions.

\textbf{UOT Loss Formulation.}
Consider a mini-batch of $N_{batch}$ source samples $\{(o^i_{src}, x^i_{src})\}_{i=1}^{N_{batch}}$ and $N_{batch}$ target samples $\{(o^j_{tgt}, x^j_{tgt})\}_{j=1}^{N_{batch}}$. Let their empirical distributions in the joint $(f_\phi(o), x)$ space be $\hat{\mu}_{src}$ and $\hat{\mu}_{tgt}$. Our UOT loss, $L_\text{UOT}(f_\phi)$, is based on the Kantorovich formulation with entropic regularization and KL-divergence penalties for marginal relaxation:
\begin{equation}
    L_\text{UOT}(f_\phi) = \min_{\Pi \in \mathbb{R}_+^{N_{batch} \times N_{batch}}} \langle \Pi, \hat{C}_\phi \rangle_F + \epsilon \cdot \Omega(\Pi) + \tau \cdot \text{KL}(\Pi \mathbf{1} || \mathbf{p}) + \tau \cdot \text{KL}(\Pi^\top \mathbf{1} || \mathbf{q}).
    \label{eq:uot_loss}
\end{equation}
Here, $\hat{C}_\phi$ is the $N_{batch} \times N_{batch}$ ground cost matrix, where each element $(\hat{C}_\phi)_{ij}$ is computed using the joint ground cost described in Sec.~\ref{ssec:jot}.  The term $\Pi$ is the transport plan; $\epsilon > 0$ is the entropic regularization strength with $\Omega(\Pi) = \sum_{i,j} \Pi_{ij} \log \Pi_{ij}$ being the entropy, facilitating efficient solution via algorithms like Sinkhorn-Knopp \cite{cuturi2013sinkhorn}; $\tau > 0$ controls the penalty for deviating from the batch marginals $\mathbf{p}$ and $\mathbf{q}$ (typically uniform); and $\text{KL}(\cdot || \cdot)$ denotes the Kullback-Leibler divergence.

\subsection{Temporally Aligned Sampling for Effective Mini-Batch Learning}
\label{ssec:sampling}

The efficacy of mini-batch OT, including UOT, hinges on presenting the solver with comparable source and target samples within each batch. For sequential robotic data, naive random sampling of individual transitions from $D_{src}$ and $D_{tgt}$ may yield pairs from different stages of tasks, leading to noisy transport plans and sub-optimal feature alignment by $f_\phi$. Increasing the minibatch size may lead to higher likelihood of sampling aligned pairs but requires more computational resources. 
%\caelan{In the limit of increasing the batch size, no longer stochastic gradient descent}

To address this, we introduce a \textit{temporally aligned sampling} strategy designed to construct mini-batches with a higher density of meaningfully corresponding state-pairs. Our strategy leverages trajectory-level similarity as a heuristic. We first quantify similarity between source trajectories $\{\xi^k_{src}\} \subset D_{src}$ and target trajectories $\{\xi^l_{tgt}\} \subset D_{tgt}$ using Dynamic Time Warping (DTW) \cite{gold2018dynamic} on their respective proprioceptive state sequences $\{x_t\}$. 
The resulting normalized DTW distance, $\bar{d}(\xi^k_{src}, \xi^l_{tgt})=d_\text{DTW}(\xi^k_{src}, \xi^l_{tgt})/\max(|\xi^k_{src}|, |\xi^l_{tgt}|)$, reflects overall behavioral similarity.  To turn these distances into sampling weights, we apply a softplus-based transformation: $w(\xi^k_{src}, \xi^l_{tgt})=1/(1+e^{10\cdot(\bar{d}(\xi^k_{src}, \xi^l_{tgt})-0.01)})$. Mini-batch construction then proceeds by (1) sampling a pair of trajectories $(\xi_{src}, \xi_{tgt})$ with probability biased towards pairs exhibiting high similarity (i.e., low DTW distance) and (2) subsequently sampling individual transition tuples $(o_{src}, x_{src}, a_{src})$ and $(o_{tgt}, x_{tgt}, a_{tgt})$ from this selected, behaviorally similar trajectory pair.

Fine-grained temporal alignment, such as sampling around DTW-matched time steps, can optionally be employed here. We describe how the UOT loss (Eq.~\ref{eq:uot_loss}) is adapted with this new sampling procedure in 
the appendix.
%Appendix~\cite{TBD}.

This two-stage process significantly increases the likelihood that source and target samples within a mini-batch share similar proprioceptive states $x$. Consequently, the UOT optimization (Eq.~\ref{eq:uot_loss}) can more effectively focus on aligning the visual latent features $f_\phi(o_{src})$ and $f_\phi(o_{tgt})$ for these relevant state-pairs. We empirically verify the importance of this sampling strategy in appendix. 

\subsection{Joint Co-Training Framework}
\label{ssec:joint}

Putting all components together, our final approach is a joint co-training framework where the visual feature encoder $f_\phi$ and the policy $\pi_\theta(a | z, x)$ are optimized concurrently. The Unbalanced Optimal Transport loss ($L_\text{UOT}$) serves as a regularization term, guiding $f_\phi$ to learn domain-invariant and action-relevant latent representations $z=f_\phi(o)$, while standard Behavior Cloning (BC) losses drive the policy learning.  The overall training objective 
%\begin{equation}
$L(f_\phi, \pi_\theta) = L_\text{BC}(f_\phi, \pi_\theta) + \lambda \cdot L_\text{UOT}(f_\phi)$
%\label{eq:joint_objective_final}
%\end{equation}
combines these components, where $L_\text{BC}$ represents the combined behavior cloning losses calculated over both source ($D_{src}$) and target ($D_{tgt}$) datasets using a standard imitation loss (e.g., MSE). 
%\caelan{Is lower case $L_{BC}$ deliberate?}
The hyper-parameter $\lambda > 0$ balances feature alignment with policy imitation. The $L_\text{UOT}(f_\phi)$ term is computed as defined in Equation~\ref{eq:uot_loss}, with mini-batches sampled with strategy described in Sec.~\ref{ssec:sampling}.
The overall training process is detailed in 
the appendix.
%Alg.~\ref{algm:learning} in Appendix~\ref{pseudocode}.

\section{Experiments}
We aim to validate the following core hypotheses. \textbf{H1}: Our method effectively learns complex manipulation tasks in both simulation and the real world. \textbf{H2}: Our method generalizes to target domains only seen in simulation. \textbf{H3}: Our method is broadly applicable to multiple observation modalities. \textbf{H4}: Scaling up simulation data coverage improves generalization performance. 
\begin{figure}[t]
    \centering
    \includegraphics[width=1.0\linewidth]{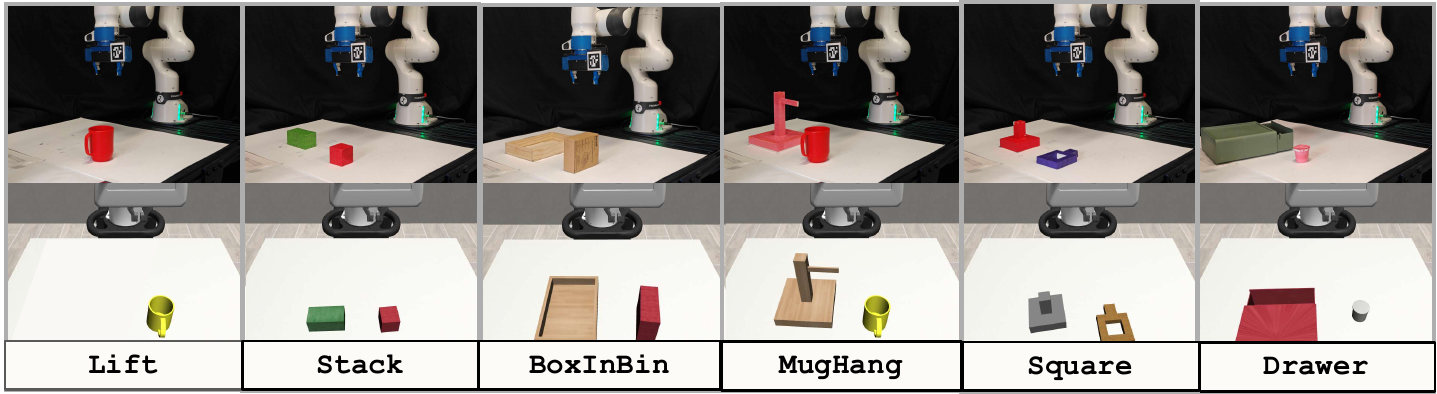}
    \caption{\textbf{Evaluation Task Suites.} We evaluate our methods on 6 different tasks in the real world ({\it top}) and simulation ({\it bottom}) to demonstrate the effectiveness of sim-to-real transfer.}
    \label{fig:tasks_list}
    \vspace{-18pt}
\end{figure}

%\ajay{Overall, the first few sections of the experiments (e.g. 5.1, 5.2, overview) are heavy in setup and technical details, making it a chore to read -- it takes a while to get to the punchlines, which are our cool results. I strongly recommend cutting as much detail as possible -- anything that is not crucial to understand our key takeaways, and move to appendix. Also, provide quantitative evidence and examples in-line wherever possible to back up your points, instead of making the reader go hunting for a table. That will greatly improve readability. Do not just say better, put a comparison in parenthesis in the text when you say better if possible.}

\subsection{Experiment Setups}

To evaluate the effectiveness of our approach, we conduct comprehensive experiments in both sim-to-sim and sim-to-real transfer scenarios on a suite of robotic tabletop manipulation tasks: \texttt{Lift}, \texttt{BoxInBin}, \texttt{Stack}, \texttt{Square}, \texttt{MugHang}, and \texttt{Drawer}. These tasks are designed to test the system’s ability to handle key challenges in robotic manipulation, including dense object interactions, long-horizon reasoning, and high-precision control.

\textbf{Environment setup.} For the real-world experiments, we deploy our method on a Franka Emika Panda robotic arm, controlled at 20Hz using joint impedance control. Visual and point cloud observations are captured using an Intel RealSense D435 depth camera.
%, and raw point cloud depth observations without colors are generated using the camera’s calibrated intrinsic and extrinsic parameters.
Our simulation environments use the Robosuite~\cite{zhu2020robosuite} simulation framework. We calibrate the camera pose and camera intrinsics in simulation to match those obtained from the real-world setup to reduce the domain gap.
%To ensure a realistic sim-to-real transfer, 

\textbf{Simulation data.} For simulation experiments, we begin by collecting 10 human demonstrations per task. Using MimicGen~\cite{mandlekar2023mimicgen}, we synthesize 200-1000 trajectories in the source domain, covering the full range of initial states (denoted as \texttt{Source}). In the target domain, we divide the reset region into two subregions: one is populated with 10 trajectories for training (denoted as \texttt{Target}), while the other remains completely held out from training (denoted as \texttt{Target-OOD}). This held-out subregion is used to evaluate each method’s generalization under Out-Of-Distribution (OOD) conditions.
% \caelan{Where do we list the number of real demonstrations?}\shuo{Appendix}

\textbf{Real data.} For real-world experiments, we adopt a similar strategy by partitioning the reset region—aligned with the simulation setup into two subregions. Based on task complexity, we collect 10–25 human demonstrations within one subregion and generate 1000 simulated trajectories. To evaluate generalization in OOD scenarios, we consider the following settings in the real world: \texttt{Shape}, where the test object has not been seen during real data collection; \texttt{Reset}, where the initial object pose falls outside the range covered by demonstrations; and \texttt{Texture}, where the object is wrapped in a novel texture not present in any real-world training data. Detailed visualizations of each task and reset configuration are included in 
the appendix.
%Appendix~\ref{sec:task_setups}.

\textbf{Observation modality and domain gaps.} We evaluate our approach using two observation modalities: point clouds and RGB images. For point cloud observations, our method and the baselines adapt 3D Diffusion Policy~\citep{Ze2024DP3} with a PointNet encoder~\citep{qi2017pointnet}. For RGB image observations, we use Diffusion Policy~\citep{chi2023diffusionpolicy} with a ResNet-18 encoder~\citep{he2016deep}.
%across all methods. 
To evaluate the generalization capabilities of different methods under visual domain shifts in simulation, we introduce several target domain variations: \texttt{Viewpoint1-Point}, \texttt{Viewpoint3-Point}, \texttt{Perturbation-Point}, \texttt{Viewpoint-Image}, and \texttt{Texture-Image}. Descriptions of each domain shift are provided in 
the appendix.
%Appendix~\ref{sec:task_setups}.

\noindent\textbf{Baselines.} We compare our method against the following baselines: \texttt{MMD}---minimizes the distance between the mean embeddings of source and target data~\citep{tzeng2014deep}; 
%\caelan{Is MMD ever defined?}
\texttt{Co-training}---trains the model using a mixed batch of source and target domain data, following the strategy proposed by~\citep{wei2025empirical, maddukuri2025sim}; \texttt{Source-only}---trains the model exclusively on data from the source domain, which in sim-to-real experiments corresponds to using only simulation data; \texttt{Target-only}---trains the model exclusively on data from the target domain, which in sim-to-real experiments trains with only real world data.

\subsection{Cross-Domain Generalization Results}

\begin{table}[h]
\centering
%\scriptsize 
\footnotesize
\setlength{\tabcolsep}{2pt}

\begin{tabular}{l|cc|cc|cc|cc|cc|cc|cc}
\specialrule{1.pt}{0pt}{0pt}   % thick line before header
 & \multicolumn{2}{c|}{\shortstack{\texttt{Stack (V)}}}
 & \multicolumn{2}{c|}{\shortstack{\texttt{Square (V)}}}
 & \multicolumn{2}{c|}{\shortstack{\texttt{BoxInBin (V)}}}
 & \multicolumn{2}{c|}{\shortstack{\texttt{Stack (T)}}}
 & \multicolumn{2}{c|}{\shortstack{\texttt{Square (T)}}}
 & \multicolumn{2}{c|}{\shortstack{\texttt{BoxInBin (T)}}}
 & \multicolumn{2}{c}{\shortstack{\textbf{Average}}}\\
 & \textbf{T} & \textbf{T‑O}
 & \textbf{T} & \textbf{T‑O}
 & \textbf{T} & \textbf{T‑O}
 & \textbf{T} & \textbf{T‑O}
 & \textbf{T} & \textbf{T‑O}
 & \textbf{T} & \textbf{T‑O}
 & \textbf{T} & \textbf{T‑O}\\

\midrule
\texttt{S.‑only} & 0.00 & 0.00 & 0.00 & 0.00 & 0.00 & 0.00 & 0.00 & 0.00 & 0.00 & 0.00 & 0.00 & 0.00 & 0.00 & 0.00 \\
\texttt{T.‑only} & 0.30 & 0.00 & 0.20 & 0.00 & 0.82 & 0.00 & 0.42 & 0.00 & 0.48 & 0.00 & 0.64 & 0.00 & 0.48 & 0.00\\
\texttt{MMD} & 0.38 & 0.00 & 0.18 & \textbf{0.04} &  0.82 & 0.16  & 0.44 & 0.4 & 0.38 & 0.34  & 0.80 & 0.70 & 0.50 & 0.30 \\
\texttt{Co‑train.} & 0.44 & \textbf{0.04} & 0.76 & 0.00  & \textbf{0.90} & 0.14 & 0.54 & 0.34 & 0.66 & 0.46 & \textbf{0.98} & 0.72 & 0.71 & 0.28 \\
\texttt{Ours} & \textbf{0.65} & \textbf{0.04} & \textbf{0.86} & 0.02 &  0.88 & \textbf{0.26} & \textbf{0.66} & \textbf{0.52} & \textbf{0.68} & \textbf{0.54} & 0.96 & \textbf{0.82} & \textbf{0.78} & \textbf{0.36}\\
\specialrule{1.pt}{0pt}{0pt}   % thick line after data
\end{tabular}

\caption{\textbf{Sim-to-Sim Success Rates for Image-Based Policies}. \texttt{V} and \texttt{T} represent \texttt{Viewpoint-Image} and \texttt{Texture-Image} domain shifts, respectively. \textbf{T} and \textbf{T-O} correspond to the target domain and target domain with out-of-distribution (OOD) scenarios. % \textbf{S}, source domain
}\label{exp:img_sim}
\vspace{-20pt}
\end{table}

% \begin{table}[h]
% \centering
% \begin{tabular}{l|cc|cc|cc|c}
% \specialrule{1.pt}{0pt}{0pt}   % thick line before header
%  & \multicolumn{2}{c|}{\shortstack{\texttt{Stack}}}
%  & \multicolumn{2}{c|}{\shortstack{\texttt{Square}}}
%  & \multicolumn{2}{c|}{\shortstack{\texttt{BoxInBin}}}
%  & \textbf{Average} \\ 
%  & grasp & full & grasp & full & grasp & full & full \\
% \midrule
% \texttt{Source-only}   & 0.1 & 0.0 & 0.0 & 0.0 & 0.0 & 0.0 & 0.00 \\
%  \texttt{Target-only}  & 0.7 & 0.7 & 0.8 & 0.0 & 0.7 & 0.7 & 0.47 \\
% \texttt{Co‑training}   & 0.8 & 0.7 & 0.8 & 0.1 & \textbf{0.9} & 0.8 & 0.53 \\
% \texttt{Ours}       & \textbf{0.9} & \textbf{0.9} & \textbf{0.9} & \textbf{0.4} & \textbf{0.9} & \textbf{0.9} & \textbf{0.73} \\
% \specialrule{1.pt}{0pt}{0pt}   % thick line after data
% \end{tabular}
% \caption{\textbf{Real World Image-Based Policy In-Distribution Success Rates.} The \textbf{Average} denotes the average full task success rates over all tasks.
% \caelan{Move to appendix}
% % \caelan{Use texttt for \texttt{co-train} and other approaches for consistency}
% }\label{exp:img_real_id}
% \end{table}

\begin{table}[h]
\centering
\footnotesize
\begin{tabular}{l|cc|cc|cc|c}
\specialrule{1.pt}{0pt}{0pt}   % thick line before header
 & \multicolumn{2}{c|}{\shortstack{\texttt{Stack (R)}}}
 & \multicolumn{2}{c|}{\shortstack{\texttt{Square (R)}}}
 & \multicolumn{2}{c|}{\shortstack{\texttt{BoxInBin (T)}}}
 & \textbf{Average} \\ 
 & grasp & full & grasp & full & grasp & full & full \\
\midrule
\texttt{Source-only}  & 0.0 & 0.0 & 0.0 & 0.0 & 0.0 & 0.0 & 0.00 \\
\texttt{Target-only}  & 0.0 & 0.0 & 0.1 & 0.0 & 0.0 & 0.0 & 0.00 \\
\texttt{Co-training}   & 0.0 & 0.0 & 0.3 & 0.0 & 0.4 & 0.3 & 0.10 \\
\texttt{Ours}       & \textbf{0.4} & \textbf{0.4} & \textbf{0.5} & \textbf{0.1} & \textbf{0.7} & \textbf{0.7} & \textbf{0.40} \\
\specialrule{1.pt}{0pt}{0pt}   % thick line after data
\end{tabular}
\caption{\textbf{Real-World Image-Based Policy OOD Success Rates.} \texttt{R} and \texttt{T} denote \texttt{Reset} OOD and \texttt{Texture} OOD, respectively. The \textbf{Average} denotes the average full task success rates over all tasks.}\label{exp:img_real_ood}
\vspace{-8pt}
\end{table}

\begin{table}[h]
\centering
\scriptsize
% \small
% \foosize 
\setlength{\tabcolsep}{5pt}

% \begin{tabular}{l|ccc|ccc|ccc|ccc|ccc|ccc}
% \specialrule{1.pt}{0pt}{0pt}   % thick line before header
%  & \multicolumn{3}{c|}{\shortstack{\texttt{BoxInBin (V3)}}}
%  & \multicolumn{3}{c|}{\shortstack{\texttt{BoxInBin (P)}}}
%  & \multicolumn{3}{c|}{\shortstack{\texttt{Lift (V1)}}}
%  & \multicolumn{3}{c|}{\shortstack{\texttt{Stack (V1)}}}
%  & \multicolumn{3}{c|}{\shortstack{\texttt{Square (V1)}}}
%  & \multicolumn{3}{c}{\shortstack{\texttt{MugHang (V1)}}} \\
%  & S & T & T‑O
%  & S & T & T‑O
%  & S & T & T‑O
%  & S & T & T‑O
%  & S & T & T‑O
%  & S & T & T‑O \\
% \midrule
% \texttt{S.‑only}   & \textbf{0.77} & 0.08 & 0.10 & 0.77 & 0.52 & 0.60 & 0.90 & 0.32 & 0.40 & 0.96 & 0.52 & 0.64 & \textbf{0.43} & 0.10 & 0.08 & 0.34 & 0.12 & 0.10 \\
% \texttt{T.‑only}   & 0.05 & 0.42 & 0.00 & 0.29 & 0.58 & 0.00 & 0.03 & 0.60 & 0.00 & 0.00 & 0.32 & 0.00 & 0.02 & 0.16 & 0.00 & 0.00 & 0.18 & 0.00 \\
% \texttt{MMD}      & 0.72 & 0.50 & 0.38 & \textbf{0.84} & 0.66 & 0.50 & 0.68 & 0.56 & 0.52 & 0.82 & 0.70 & 0.66 & 0.18 & 0.18 & 0.12 & 0.20 & 0.18 & 0.20 \\
% \texttt{Co‑train.} & 0.75 & 0.76 & 0.52 & 0.71 & 0.70 & 0.66 & \textbf{0.83} & \textbf{0.92} & 0.48 & 0.92 & \textbf{0.86} & 0.72 & 0.31 & 0.24 & 0.24 & 0.48 & 0.26 & 0.22 \\
% \texttt{Ours}     & 0.72 & \textbf{0.84} & \textbf{0.58} & 0.80 & \textbf{0.80} & \textbf{0.76} & 0.78 & 0.80 & \textbf{0.60} & \textbf{0.97} & 0.82 & \textbf{0.86} & 0.34 & \textbf{0.42} & \textbf{0.38} & \textbf{0.51} & \textbf{0.40} & \textbf{0.34} \\
% \specialrule{1.pt}{0pt}{0pt}   % thick line after data
% \end{tabular}

\begin{tabular}{l|cc|cc|cc|cc|cc|cc|cc}
\specialrule{1.pt}{0pt}{0pt}   % thick line before header
 & \multicolumn{2}{c|}{\shortstack{\texttt{BoxInBin (V3)}}}
 & \multicolumn{2}{c|}{\shortstack{\texttt{BoxInBin (P)}}}
 & \multicolumn{2}{c|}{\shortstack{\texttt{Lift (V1)}}}
 & \multicolumn{2}{c|}{\shortstack{\texttt{Stack (V1)}}}
 & \multicolumn{2}{c|}{\shortstack{\texttt{Square (V1)}}}
 & \multicolumn{2}{c|}{\shortstack{\texttt{MugHang (V1)}}}
 & \multicolumn{2}{c}{\shortstack{\textbf{Average}}} \\
 & T & T‑O
 & T & T‑O
 & T & T‑O
 & T & T‑O
 & T & T‑O
 & T & T‑O
 & T & T‑O\\
\midrule
\texttt{S.‑only} & 0.08 & 0.10 & 0.52 & 0.60 & 0.32 & 0.40 & 0.52 & 0.64 & 0.10 & 0.08 & 0.12 & 0.10 & 0.28 & 0.32 \\
\texttt{T.‑only} & 0.42 & 0.00 & 0.58 & 0.00 & 0.60 & 0.00 & 0.32 & 0.00 & 0.16 & 0.00 & 0.18 & 0.00 & 0.38 & 0.00 \\
\texttt{MMD} & 0.50 & 0.38 & 0.66 & 0.50 & 0.56 & 0.52 & 0.70 & 0.66 & 0.18 & 0.12 & 0.18 & 0.20 & 0.46 & 0.40 \\
\texttt{Co‑train.} & 0.76 & 0.52 & 0.70 & 0.66 & \textbf{0.92} & 0.48 & \textbf{0.86} & 0.72 & 0.24 & 0.24 & 0.26 & 0.22 & 0.62 & 0.47 \\
\texttt{Ours} & \textbf{0.84} & \textbf{0.58} & \textbf{0.80} & \textbf{0.76} & 0.80 & \textbf{0.60} & 0.82 & \textbf{0.86} & \textbf{0.42} & \textbf{0.38} & \textbf{0.40} & \textbf{0.34} & \textbf{0.68} & \textbf{0.59} \\
\specialrule{1.pt}{0pt}{0pt}   % thick line after data
\end{tabular}

\caption{\textbf{Sim-to-sim Success Rates For Point Cloud-Based Policies.} \texttt{V1}, \texttt{V3}, and \texttt{P} indicate domain shifts due to \texttt{Viewpoint1-Point}, \texttt{Viewpoint3-Point}, and \texttt{Perturbation-Point}, respectively. \textbf{T} and \textbf{T-O} denote the target domain and target domain under out-of-distribution (OOD) conditions. % \textbf{S} source domain
}\label{exp:pc_sim}
\vspace{-8pt}
\end{table}

\begin{table}[h]
\centering
% \scriptsize 
\footnotesize
\setlength{\tabcolsep}{3pt} 
\begin{tabular}{l|cc|cc|cc|cc|cc|cc|c}
\specialrule{1.pt}{0pt}{0pt}   % thick line before header
 & \multicolumn{2}{c|}{\shortstack{\texttt{Stack (R)}}} 
 & \multicolumn{2}{c|}{\shortstack{\texttt{Square (R)}}}
 & \multicolumn{2}{c|}{\shortstack{\texttt{BoxInBin (R)}}}
 & \multicolumn{2}{c|}{\shortstack{\texttt{Lift (R)}}}
 & \multicolumn{2}{c|}{\shortstack{\texttt{Lift (S)}}}
 & \multicolumn{2}{c|}{\shortstack{\texttt{Lift (R+S)}}}
 & \textbf{Average} \\
 & grasp & full & grasp & full & grasp & full & reach & full & reach & full & reach & full & full \\
\midrule
\texttt{S.-only}   & 0.6 & 0.3 & 0.1 & 0.1 & 0.3 & 0.2 & 0.8 & 0.8 & 0.6 & 0.6 & 0.9 & 0.9 & 0.48 \\
\texttt{T.-only}  & 0.0 & 0.0 & 0.3 & 0.0 & 0.0 & 0.0 & 0.0 & 0.0 & \textbf{1.0} & \textbf{1.0} & 0.0 & 0.0 & 0.17 \\
\texttt{Co-train.}   & 0.4 & 0.1 & \textbf{0.6} & \textbf{0.2} & 0.2 & 0.0 & 0.8 & 0.8 & \textbf{1.0} & \textbf{1.0} & 0.9 & 0.9 & 0.50 \\
\texttt{Ours}       & \textbf{0.8} & \textbf{0.4} & \textbf{0.6} & 0.1 & \textbf{0.7} & \textbf{0.5} & \textbf{1.0} & \textbf{1.0} & \textbf{1.0} & \textbf{1.0} & \textbf{1.0} & \textbf{1.0} & \textbf{0.67} \\
\specialrule{1.pt}{0pt}{0pt}   % thick line after data
\end{tabular}
\caption{\textbf{Real World Point-Cloud-Based Policy OOD Success Rates.} \texttt{R} and \texttt{S} denote \texttt{Reset} OOD and \texttt{Shape} OOD, respectively.  The \textbf{Average} denotes the average full task success rates over all tasks.}\label{exp:pc_real_ood}
\vspace{-18pt}
\end{table}

We report results for policies using point cloud and image-based observations in both simulation and real-world settings. For simulation, image based and point cloud based performance are shown in Tables~\ref{exp:img_sim} and~\ref{exp:pc_sim}. For real-world experiments, in-distribution results are presented in Tables~\ref{exp:img_real_id} and~\ref{exp:pc_real_id} in Appendix, while out-of-distribution (OOD) performance is reported in Tables~\ref{exp:img_real_ood} and~\ref{exp:pc_real_ood}. 

\textbf{These results support the following key hypotheses:}

\textbf{Our method effectively learns complex manipulation tasks in both simulation and the real world (H1).} Experimental results show that our approach consistently matches or outperforms in terms of success rates all baselines across source and target domains in both simulated and real-world settings. On real-world tasks, our method achieves average success rates of 0.73 and 0.77 for image-based and point cloud-based policies, respectively.
% The \texttt{Source-only} and 
The \texttt{Target-only} baseline performs well 
in distribution
%in the scenarios that covered with training data, 
but fails to generalize under domain shifts. The \texttt{MMD} baseline~\cite{tzeng2014deep} offers limited improvement by aligning global feature statistics, but its coarse alignment often disrupts task-relevant structure and harms source-domain performance.
In contrast, by using Unbalanced Optimal Transport, our method performs selective, structure-aware alignment, avoiding spurious matches. 
%By incorporating proprioception as an auxiliary cue, it further refines alignment toward behaviorally consistent states, leading to better generalization without sacrificing performance in the source domain.

\textbf{Our method generalizes to target domains only seen in simulation (H2).} In \texttt{Target-OOD} scenarios, our method outperforms all baselines, underscoring the value of learning domain-invariant representations (see Tab.~\ref{exp:img_real_ood} and Tab.~\ref{exp:pc_real_ood}). While \texttt{Co-training} baselines~\cite{wei2025empirical, maddukuri2025sim} perform well when the target-domain training data overlaps with the evaluation region, they struggle to generalize when this overlap is absent. This limitation is especially evident under large domain shifts. For example, in the real-world \texttt{BoxInBin} and \texttt{Stack} tasks with novel textures or reset poses, our method achieves success rates of 0.7 and 0.4, respectively, using image-based observations—compared to just 0.3 and 0.0 for the \texttt{Co-training} baseline. These results highlight the shortcomings of relying purely on supervised target-domain data without explicitly addressing domain shift.

\textbf{Our method is broadly applicable to multiple observation modalities (H3).}
We observe consistent performance gains across both image-based and point cloud inputs. In simulation, policies trained with either modality outperform all baselines, demonstrating that our approach effectively learns domain-invariant features that capture task-relevant information on multiple sensory modalities. 

\textbf{Simulation data provides a scalable and effective way to augment real-world training.}
Across real-world tasks, both our method and the \texttt{Co-training} baseline benefit significantly from augmenting limited real-world demonstrations with simulated data. Policies trained with this augmented data consistently outperform the \texttt{Target-only} baseline, especially in out-of-distribution (OOD) settings where real-world coverage is sparse. This underscores the value of using low-cost simulation data to fill in gaps in real-world datasets, enabling more scalable and generalizable behavior cloning.

\begin{figure}[h]
    \centering
    \includegraphics[width=\linewidth]{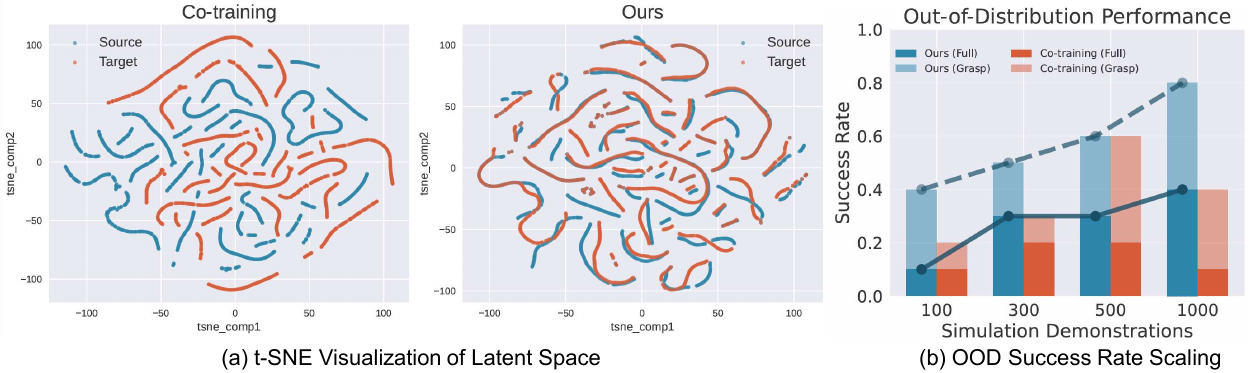}
    \caption{(a) \textbf{Latent Space Visualization.} Our OT alignment maps source domain samples (blue) and target domain samples (red) nearby in the latent space, yielding a single, well‑mixed cluster. This overlap demonstrates that OT alignment effectively synchronizes cross‑domain feature distributions, improving sim-to-real transfer. (b) \textbf{Out-Of-Distribution Performance.} Scaling the number of simulation demonstrations leads to significant OOD success rate gains. 
    %Importantly, our approach benefits more from additional data and shows a markedly stronger scaling trend compared to co‑training.
    }
    % \caelan{Since the in distribution performance is not super interesting and the figures are small, maybe we omit the In-Distribution.}}
    % \caelan{Performance is too generic. Say success rate instead.}
    \label{fig:sim-data-scaling-and-tsne}
    \vspace{-10pt}
\end{figure}

\textbf{Scaling up simulation data coverage improves real-world performance (H4).} To analyze simulation data scaling, we consider the \texttt{Stack} task in the real world with point cloud observation. We generated 100, 300, 500, and 1000 simulated trajectories, combined them with 25 real-world demonstrations, and trained both our method and the \texttt{Co-training} baseline. As shown in Fig.~\ref{fig:sim-data-scaling-and-tsne}(b), increasing the amount of simulation data significantly improves our method's performance in target domain regions that lack real-world coverage, highlighting the importance of learning a domain-invariant latent space that enables the policy to generalize beyond observed distributions.
%thereby allowing low-cost simulation data to be effectively leveraged for scalable and robust behavior cloning.

\textbf{Our method learns shared latent space that aligns simulation and real data.} To better understand how the learned embeddings contribute to generalization, we visualize them using t-SNE~\cite{van2008visualizing}, as shown in Fig.~\ref{fig:sim-data-scaling-and-tsne}(a). Blue and red correspond to features extracted from source and target domain observations, respectively. The left plot shows embeddings from the encoder trained with the \texttt{Co-training} baseline, while the right plot shows embeddings from our method. The visualization reveals that our approach leads to significantly better alignment between source and target distributions, highlighting its ability to learn domain-invariant representations that facilitate robust generalization. We also visualize the transport plan on a randomly sampled batch, along with the corresponding image observations from the source and target domains. As shown in Fig.~\ref{fig:transport-plan} in the Appendix, the transport plan effectively aligns data points with similar states across domains. % \danfei{Mention transport map here too}

\section{Limitations and Conclusions}

The main limitation of our work is that we only address sim-to-real visual observation gaps. 
% Sim and real demonstration alignment
Addressing action dynamics gaps remains future work.
Because we use MimicGen~\cite{mandlekar2023mimicgen} for automated simulation demonstration generation, we inherit its limitations, namely, our method primarily applies to tasks with prehensile and quasi-static interactions.
We require a small number of real-world on-task demonstrations that are aligned with the simulated demonstrations. Future work involves relaxing this requirement, for example, by instead consuming unstructured real-world data, such as play data.

In conclusion, we presented a framework for effectively incorporating large datasets of simulation demonstrations into real-world policy learning pipelines via feature-consistent co-training. 
We proposed using Optimal Transport (OT) to align encoder features to be invariant to whether observations are from simulation or the real world, improving the transferability of simulation data.
Because we have much more simulation data than real-world data, we incorporated an Unbalanced OT loss within our training objective and devised a data sampling scheme that explicitly yields similar simulation and real-world demonstration pairs.
Finally, we demonstrated the improved learning performance arising from the simulated demonstrations both in a sim-to-sim testbed as well as in real-world tasks.

% In Sec.~\ref{ssec:jot}, we assume the robot's proprioceptive state suffices as a surrogate for actions during alignment; however, this may not hold in tasks with forceful interactions.
% Similarly, in Sec.~\ref{ssec:sampling}, use apply DTW on the proprioceptive state sequence, which lacks information about contact, which may affect task completion.
%\danfei{This is mandated by the checklist}
% The main limitation of our method is the assumption that the simulation and real world demonstration data have aligned behaviors. While this is achievable with data generation tools such as MimicGen, the assumption becomes harder to satisfy in settings such as multitask learning. In future works, we seek to adopt less restrictive alignment guidance such as natural language description obtained from Vision-Language models and visual features extracted from foundation models, which can improve the applicability of our method.
%\caelan{Overlap with the limitations section now}

\section{Acknowledgment}
This work was supported in part by the National Science Foundation under Awards No. 2409016 and 2442393. Any opinions, findings, and conclusions or recommendations expressed in this material are those of the author(s) and do not necessarily reflect the views of the National Science Foundation.
%%%%%%%%%%%%%%%%%%%%%%%%%%%%%%%%%%%%%%%%%%%%%%%%%%%%%%%%%%%%

%%%%%%%%%%%%%%%%%%%%%%%%%%%%%%%%%%%%%%%%%%%%%%%%%%%%%%%%%%%%

% \newpage
% \input{text/checklist}

\newpage
\bibliographystyle{unsrt}
\bibliography{neurips_2025}

@article{deitke2022️,
    title={ProcTHOR: Large-Scale Embodied AI Using Procedural Generation},
    author={Deitke, Matt and VanderBilt, Eli and Herrasti, Alvaro and Weihs, Luca and Ehsani, Kiana and Salvador, Jordi and Han, Winson and Kolve, Eric and Kembhavi, Aniruddha and Mottaghi, Roozbeh},
    journal={Advances in Neural Information Processing Systems},
    volume={35},
    pages={5982--5994},
    year={2022}
}

@inproceedings{
    mandlekar2023mimicgen,
    title={MimicGen: A Data Generation System for Scalable Robot Learning using Human Demonstrations},
    author={Ajay Mandlekar and Soroush Nasiriany and Bowen Wen and Iretiayo Akinola and Yashraj Narang and Linxi Fan and Yuke Zhu and Dieter Fox},
    booktitle={7th Annual Conference on Robot Learning},
    year={2023},
    url={https://openreview.net/forum?id=dk-2R1f_LR}
}

@article{jiang2024dexmimicgen,
  title={Dexmimicgen: Automated data generation for bimanual dexterous manipulation via imitation learning},
  author={Jiang, Zhenyu and Xie, Yuqi and Lin, Kevin and Xu, Zhenjia and Wan, Weikang and Mandlekar, Ajay and Fan, Linxi and Zhu, Yuke},
  journal={arXiv preprint arXiv:2410.24185},
  year={2024}
}

@article{cheng2023nod,
    title={NOD-TAMP: Multi-Step Manipulation Planning with Neural Object Descriptors},
    author={Cheng, Shuo and Garrett, Caelan and Mandlekar, Ajay and Xu, Danfei},
    journal={arXiv preprint arXiv:2311.01530},
    year={2023}
}

@inproceedings{
    cheng2024nodtamp,
    title={{NOD}-{TAMP}: Generalizable Long-Horizon Planning with Neural Object Descriptors},
    author={Shuo Cheng and Caelan Reed Garrett and Ajay Mandlekar and Danfei Xu},
    booktitle={8th Annual Conference on Robot Learning},
    year={2024},
    url={https://openreview.net/forum?id=rThtgkXuvZ}
}

@article{chi2023diffusion,
  title={Diffusion policy: Visuomotor policy learning via action diffusion},
  author={Chi, Cheng and Xu, Zhenjia and Feng, Siyuan and Cousineau, Eric and Du, Yilun and Burchfiel, Benjamin and Tedrake, Russ and Song, Shuran},
  journal={The International Journal of Robotics Research},
  pages={02783649241273668},
  year={2023},
  publisher={SAGE Publications Sage UK: London, England}
}

@inproceedings{Ze2024DP3,
    title={3D Diffusion Policy: Generalizable Visuomotor Policy Learning via Simple 3D Representations},
    author={Yanjie Ze and Gu Zhang and Kangning Zhang and Chenyuan Hu and Muhan Wang and Huazhe Xu},
    booktitle={Proceedings of Robotics: Science and Systems (RSS)},
    year={2024}
}

@software{Genesis,
  author = {Genesis Authors},
  title = {Genesis: A Universal and Generative Physics Engine for Robotics and Beyond},
  month = {December},
  year = {2024},
  url = {https://github.com/Genesis-Embodied-AI/Genesis}
}

@article{cuturi2013sinkhorn,
  title={Sinkhorn distances: Lightspeed computation of optimal transport},
  author={Cuturi, Marco},
  journal={Advances in neural information processing systems},
  volume={26},
  year={2013}
}

@article{chizat2018scaling,
  title={Scaling algorithms for unbalanced optimal transport problems},
  author={Chizat, Lenaic and Peyr{\'e}, Gabriel and Schmitzer, Bernhard and Vialard, Fran{\c{c}}ois-Xavier},
  journal={Mathematics of computation},
  volume={87},
  number={314},
  pages={2563--2609},
  year={2018}
}

@article{khazatsky2024droid,
    title   = {DROID: A Large-Scale In-The-Wild Robot Manipulation Dataset},
    author  = {Alexander Khazatsky and Karl Pertsch and Suraj Nair and Ashwin Balakrishna and Sudeep Dasari and Siddharth Karamcheti and Soroush Nasiriany and Mohan Kumar Srirama and Lawrence Yunliang Chen and Kirsty Ellis and Peter David Fagan and Joey Hejna and Masha Itkina and Marion Lepert and Yecheng Jason Ma and Patrick Tree Miller and Jimmy Wu and Suneel Belkhale and Shivin Dass and Huy Ha and Arhan Jain and Abraham Lee and Youngwoon Lee and Marius Memmel and Sungjae Park and Ilija Radosavovic and Kaiyuan Wang and Albert Zhan and Kevin Black and Cheng Chi and Kyle Beltran Hatch and Shan Lin and Jingpei Lu and Jean Mercat and Abdul Rehman and Pannag R Sanketi and Archit Sharma and Cody Simpson and Quan Vuong and Homer Rich Walke and Blake Wulfe and Ted Xiao and Jonathan Heewon Yang and Arefeh Yavary and Tony Z. Zhao and Christopher Agia and Rohan Baijal and Mateo Guaman Castro and Daphne Chen and Qiuyu Chen and Trinity Chung and Jaimyn Drake and Ethan Paul Foster and Jensen Gao and David Antonio Herrera and Minho Heo and Kyle Hsu and Jiaheng Hu and Donovon Jackson and Charlotte Le and Yunshuang Li and Kevin Lin and Roy Lin and Zehan Ma and Abhiram Maddukuri and Suvir Mirchandani and Daniel Morton and Tony Nguyen and Abigail O'Neill and Rosario Scalise and Derick Seale and Victor Son and Stephen Tian and Emi Tran and Andrew E. Wang and Yilin Wu and Annie Xie and Jingyun Yang and Patrick Yin and Yunchu Zhang and Osbert Bastani and Glen Berseth and Jeannette Bohg and Ken Goldberg and Abhinav Gupta and Abhishek Gupta and Dinesh Jayaraman and Joseph J Lim and Jitendra Malik and Roberto Martín-Martín and Subramanian Ramamoorthy and Dorsa Sadigh and Shuran Song and Jiajun Wu and Michael C. Yip and Yuke Zhu and Thomas Kollar and Sergey Levine and Chelsea Finn},
    year    = {2024},
}

@misc{open_x_embodiment_rt_x_2023,
title={Open {X-E}mbodiment: Robotic Learning Datasets and {RT-X} Models},
author = {Open X-Embodiment Collaboration and Abby O'Neill and Abdul Rehman and Abhinav Gupta and Abhiram Maddukuri and Abhishek Gupta and Abhishek Padalkar and Abraham Lee and Acorn Pooley and Agrim Gupta and Ajay Mandlekar and Ajinkya Jain and Albert Tung and Alex Bewley and Alex Herzog and Alex Irpan and Alexander Khazatsky and Anant Rai and Anchit Gupta and Andrew Wang and Andrey Kolobov and Anikait Singh and Animesh Garg and Aniruddha Kembhavi and Annie Xie and Anthony Brohan and Antonin Raffin and Archit Sharma and Arefeh Yavary and Arhan Jain and Ashwin Balakrishna and Ayzaan Wahid and Ben Burgess-Limerick and Beomjoon Kim and Bernhard Schölkopf and Blake Wulfe and Brian Ichter and Cewu Lu and Charles Xu and Charlotte Le and Chelsea Finn and Chen Wang and Chenfeng Xu and Cheng Chi and Chenguang Huang and Christine Chan and Christopher Agia and Chuer Pan and Chuyuan Fu and Coline Devin and Danfei Xu and Daniel Morton and Danny Driess and Daphne Chen and Deepak Pathak and Dhruv Shah and Dieter Büchler and Dinesh Jayaraman and Dmitry Kalashnikov and Dorsa Sadigh and Edward Johns and Ethan Foster and Fangchen Liu and Federico Ceola and Fei Xia and Feiyu Zhao and Felipe Vieira Frujeri and Freek Stulp and Gaoyue Zhou and Gaurav S. Sukhatme and Gautam Salhotra and Ge Yan and Gilbert Feng and Giulio Schiavi and Glen Berseth and Gregory Kahn and Guangwen Yang and Guanzhi Wang and Hao Su and Hao-Shu Fang and Haochen Shi and Henghui Bao and Heni Ben Amor and Henrik I Christensen and Hiroki Furuta and Homanga Bharadhwaj and Homer Walke and Hongjie Fang and Huy Ha and Igor Mordatch and Ilija Radosavovic and Isabel Leal and Jacky Liang and Jad Abou-Chakra and Jaehyung Kim and Jaimyn Drake and Jan Peters and Jan Schneider and Jasmine Hsu and Jay Vakil and Jeannette Bohg and Jeffrey Bingham and Jeffrey Wu and Jensen Gao and Jiaheng Hu and Jiajun Wu and Jialin Wu and Jiankai Sun and Jianlan Luo and Jiayuan Gu and Jie Tan and Jihoon Oh and Jimmy Wu and Jingpei Lu and Jingyun Yang and Jitendra Malik and João Silvério and Joey Hejna and Jonathan Booher and Jonathan Tompson and Jonathan Yang and Jordi Salvador and Joseph J. Lim and Junhyek Han and Kaiyuan Wang and Kanishka Rao and Karl Pertsch and Karol Hausman and Keegan Go and Keerthana Gopalakrishnan and Ken Goldberg and Kendra Byrne and Kenneth Oslund and Kento Kawaharazuka and Kevin Black and Kevin Lin and Kevin Zhang and Kiana Ehsani and Kiran Lekkala and Kirsty Ellis and Krishan Rana and Krishnan Srinivasan and Kuan Fang and Kunal Pratap Singh and Kuo-Hao Zeng and Kyle Hatch and Kyle Hsu and Laurent Itti and Lawrence Yunliang Chen and Lerrel Pinto and Li Fei-Fei and Liam Tan and Linxi "Jim" Fan and Lionel Ott and Lisa Lee and Luca Weihs and Magnum Chen and Marion Lepert and Marius Memmel and Masayoshi Tomizuka and Masha Itkina and Mateo Guaman Castro and Max Spero and Maximilian Du and Michael Ahn and Michael C. Yip and Mingtong Zhang and Mingyu Ding and Minho Heo and Mohan Kumar Srirama and Mohit Sharma and Moo Jin Kim and Naoaki Kanazawa and Nicklas Hansen and Nicolas Heess and Nikhil J Joshi and Niko Suenderhauf and Ning Liu and Norman Di Palo and Nur Muhammad Mahi Shafiullah and Oier Mees and Oliver Kroemer and Osbert Bastani and Pannag R Sanketi and Patrick "Tree" Miller and Patrick Yin and Paul Wohlhart and Peng Xu and Peter David Fagan and Peter Mitrano and Pierre Sermanet and Pieter Abbeel and Priya Sundaresan and Qiuyu Chen and Quan Vuong and Rafael Rafailov and Ran Tian and Ria Doshi and Roberto Mart{'i}n-Mart{'i}n and Rohan Baijal and Rosario Scalise and Rose Hendrix and Roy Lin and Runjia Qian and Ruohan Zhang and Russell Mendonca and Rutav Shah and Ryan Hoque and Ryan Julian and Samuel Bustamante and Sean Kirmani and Sergey Levine and Shan Lin and Sherry Moore and Shikhar Bahl and Shivin Dass and Shubham Sonawani and Shubham Tulsiani and Shuran Song and Sichun Xu and Siddhant Haldar and Siddharth Karamcheti and Simeon Adebola and Simon Guist and Soroush Nasiriany and Stefan Schaal and Stefan Welker and Stephen Tian and Subramanian Ramamoorthy and Sudeep Dasari and Suneel Belkhale and Sungjae Park and Suraj Nair and Suvir Mirchandani and Takayuki Osa and Tanmay Gupta and Tatsuya Harada and Tatsuya Matsushima and Ted Xiao and Thomas Kollar and Tianhe Yu and Tianli Ding and Todor Davchev and Tony Z. Zhao and Travis Armstrong and Trevor Darrell and Trinity Chung and Vidhi Jain and Vikash Kumar and Vincent Vanhoucke and Wei Zhan and Wenxuan Zhou and Wolfram Burgard and Xi Chen and Xiangyu Chen and Xiaolong Wang and Xinghao Zhu and Xinyang Geng and Xiyuan Liu and Xu Liangwei and Xuanlin Li and Yansong Pang and Yao Lu and Yecheng Jason Ma and Yejin Kim and Yevgen Chebotar and Yifan Zhou and Yifeng Zhu and Yilin Wu and Ying Xu and Yixuan Wang and Yonatan Bisk and Yongqiang Dou and Yoonyoung Cho and Youngwoon Lee and Yuchen Cui and Yue Cao and Yueh-Hua Wu and Yujin Tang and Yuke Zhu and Yunchu Zhang and Yunfan Jiang and Yunshuang Li and Yunzhu Li and Yusuke Iwasawa and Yutaka Matsuo and Zehan Ma and Zhuo Xu and Zichen Jeff Cui and Zichen Zhang and Zipeng Fu and Zipeng Lin},
howpublished  = {\url{https://arxiv.org/abs/2310.08864}},
year = {2023},
}

@article{pomerleau1988alvinn,
  title={Alvinn: An autonomous land vehicle in a neural network},
  author={Pomerleau, Dean A},
  journal={Advances in neural information processing systems},
  volume={1},
  year={1988}
}

@inproceedings{florence2022implicit,
  title={Implicit behavioral cloning},
  author={Florence, Pete and Lynch, Corey and Zeng, Andy and Ramirez, Oscar A and Wahid, Ayzaan and Downs, Laura and Wong, Adrian and Lee, Johnny and Mordatch, Igor and Tompson, Jonathan},
  booktitle={Conference on robot learning},
  pages={158--168},
  year={2022},
  organization={PMLR}
}

@InProceedings{Xiang_2020_SAPIEN,
author = {Xiang, Fanbo and Qin, Yuzhe and Mo, Kaichun and Xia, Yikuan and Zhu, Hao and Liu, Fangchen and Liu, Minghua and Jiang, Hanxiao and Yuan, Yifu and Wang, He and Yi, Li and Chang, Angel X. and Guibas, Leonidas J. and Su, Hao},
title = {{SAPIEN}: A SimulAted Part-based Interactive ENvironment},
booktitle = {The IEEE Conference on Computer Vision and Pattern Recognition (CVPR)},
month = {June},
year = {2020}}

@inproceedings{raistrick2024infinigen,
  title={Infinigen indoors: Photorealistic indoor scenes using procedural generation},
  author={Raistrick, Alexander and Mei, Lingjie and Kayan, Karhan and Yan, David and Zuo, Yiming and Han, Beining and Wen, Hongyu and Parakh, Meenal and Alexandropoulos, Stamatis and Lipson, Lahav and others},
  booktitle={Proceedings of the IEEE/CVF Conference on Computer Vision and Pattern Recognition},
  pages={21783--21794},
  year={2024}
}

@article{andrychowicz2020learning,
  title={Learning dexterous in-hand manipulation},
  author={Andrychowicz, OpenAI: Marcin and Baker, Bowen and Chociej, Maciek and Jozefowicz, Rafal and McGrew, Bob and Pachocki, Jakub and Petron, Arthur and Plappert, Matthias and Powell, Glenn and Ray, Alex and others},
  journal={The International Journal of Robotics Research},
  volume={39},
  number={1},
  pages={3--20},
  year={2020},
  publisher={SAGE Publications Sage UK: London, England}
}

@inproceedings{matas2018sim,
  title={Sim-to-real reinforcement learning for deformable object manipulation},
  author={Matas, Jan and James, Stephen and Davison, Andrew J},
  booktitle={Conference on Robot Learning},
  pages={734--743},
  year={2018},
  organization={PMLR}
}

@inproceedings{tobin2017domain,
  title={Domain randomization for transferring deep neural networks from simulation to the real world},
  author={Tobin, Josh and Fong, Rachel and Ray, Alex and Schneider, Jonas and Zaremba, Wojciech and Abbeel, Pieter},
  booktitle={2017 IEEE/RSJ international conference on intelligent robots and systems (IROS)},
  pages={23--30},
  year={2017},
  organization={IEEE}
}

@article{hansen2020self,
  title={Self-supervised policy adaptation during deployment},
  author={Hansen, Nicklas and Jangir, Rishabh and Sun, Yu and Aleny{\`a}, Guillem and Abbeel, Pieter and Efros, Alexei A and Pinto, Lerrel and Wang, Xiaolong},
  journal={arXiv preprint arXiv:2007.04309},
  year={2020}
}

@article{yarats2021mastering,
  title={Mastering visual continuous control: Improved data-augmented reinforcement learning},
  author={Yarats, Denis and Fergus, Rob and Lazaric, Alessandro and Pinto, Lerrel},
  journal={arXiv preprint arXiv:2107.09645},
  year={2021}
}

@inproceedings{bousmalis2017unsupervised,
  title={Unsupervised pixel-level domain adaptation with generative adversarial networks},
  author={Bousmalis, Konstantinos and Silberman, Nathan and Dohan, David and Erhan, Dumitru and Krishnan, Dilip},
  booktitle={Proceedings of the IEEE conference on computer vision and pattern recognition},
  pages={3722--3731},
  year={2017}
}

@inproceedings{ho2021retinagan,
  title={Retinagan: An object-aware approach to sim-to-real transfer},
  author={Ho, Daniel and Rao, Kanishka and Xu, Zhuo and Jang, Eric and Khansari, Mohi and Bai, Yunfei},
  booktitle={2021 IEEE International Conference on Robotics and Automation (ICRA)},
  pages={10920--10926},
  year={2021},
  organization={IEEE}
}

@article{courty2016optimal,
  title={Optimal transport for domain adaptation},
  author={Courty, Nicolas and Flamary, R{\'e}mi and Tuia, Devis and Rakotomamonjy, Alain},
  journal={IEEE transactions on pattern analysis and machine intelligence},
  volume={39},
  number={9},
  pages={1853--1865},
  year={2016},
  publisher={IEEE}
}

@article{zhao2023learning,
  title={Learning fine-grained bimanual manipulation with low-cost hardware},
  author={Zhao, Tony Z and Kumar, Vikash and Levine, Sergey and Finn, Chelsea},
  journal={arXiv preprint arXiv:2304.13705},
  year={2023}
}

@inproceedings{wu2024gello,
  title={Gello: A general, low-cost, and intuitive teleoperation framework for robot manipulators},
  author={Wu, Philipp and Shentu, Yide and Yi, Zhongke and Lin, Xingyu and Abbeel, Pieter},
  booktitle={2024 IEEE/RSJ International Conference on Intelligent Robots and Systems (IROS)},
  pages={12156--12163},
  year={2024},
  organization={IEEE}
}

@article{chi2024universal,
  title={Universal manipulation interface: In-the-wild robot teaching without in-the-wild robots},
  author={Chi, Cheng and Xu, Zhenjia and Pan, Chuer and Cousineau, Eric and Burchfiel, Benjamin and Feng, Siyuan and Tedrake, Russ and Song, Shuran},
  journal={arXiv preprint arXiv:2402.10329},
  year={2024}
}

@article{kareer2024egomimic,
  title={Egomimic: Scaling imitation learning via egocentric video},
  author={Kareer, Simar and Patel, Dhruv and Punamiya, Ryan and Mathur, Pranay and Cheng, Shuo and Wang, Chen and Hoffman, Judy and Xu, Danfei},
  journal={arXiv preprint arXiv:2410.24221},
  year={2024}
}

@inproceedings{james2017transferring,
  title={Transferring end-to-end visuomotor control from simulation to real world for a multi-stage task},
  author={James, Stephen and Davison, Andrew J and Johns, Edward},
  booktitle={Conference on Robot Learning},
  pages={334--343},
  year={2017},
  organization={PMLR}
}

@article{yuan2024learning,
  title={Learning to manipulate anywhere: A visual generalizable framework for reinforcement learning},
  author={Yuan, Zhecheng and Wei, Tianming and Cheng, Shuiqi and Zhang, Gu and Chen, Yuanpei and Xu, Huazhe},
  journal={arXiv preprint arXiv:2407.15815},
  year={2024}
}

@article{farahani2021brief,
  title={A brief review of domain adaptation},
  author={Farahani, Abolfazl and Voghoei, Sahar and Rasheed, Khaled and Arabnia, Hamid R},
  journal={Advances in data science and information engineering: proceedings from ICDATA 2020 and IKE 2020},
  pages={877--894},
  year={2021},
  publisher={Springer}
}

@inproceedings{fatras2021unbalanced,
  title={Unbalanced minibatch optimal transport; applications to domain adaptation},
  author={Fatras, Kilian and S{\'e}journ{\'e}, Thibault and Flamary, R{\'e}mi and Courty, Nicolas},
  booktitle={International Conference on Machine Learning},
  pages={3186--3197},
  year={2021},
  organization={PMLR}
}

@article{courty2017joint,
  title={Joint distribution optimal transportation for domain adaptation},
  author={Courty, Nicolas and Flamary, R{\'e}mi and Habrard, Amaury and Rakotomamonjy, Alain},
  journal={Advances in neural information processing systems},
  volume={30},
  year={2017}
}

@inproceedings{damodaran2018deepjdot,
  title={Deepjdot: Deep joint distribution optimal transport for unsupervised domain adaptation},
  author={Damodaran, Bharath Bhushan and Kellenberger, Benjamin and Flamary, R{\'e}mi and Tuia, Devis and Courty, Nicolas},
  booktitle={Proceedings of the European conference on computer vision (ECCV)},
  pages={447--463},
  year={2018}
}

@article{tzeng2014deep,
  title={Deep domain confusion: Maximizing for domain invariance},
  author={Tzeng, Eric and Hoffman, Judy and Zhang, Ning and Saenko, Kate and Darrell, Trevor},
  journal={arXiv preprint arXiv:1412.3474},
  year={2014}
}

@article{wei2025empirical,
  title={Empirical Analysis of Sim-and-Real Cotraining Of Diffusion Policies For Planar Pushing from Pixels},
  author={Wei, Adam and Agarwal, Abhinav and Chen, Boyuan and Bosworth, Rohan and Pfaff, Nicholas and Tedrake, Russ},
  journal={arXiv preprint arXiv:2503.22634},
  year={2025}
}

@article{maddukuri2025sim,
  title={Sim-and-Real Co-Training: A Simple Recipe for Vision-Based Robotic Manipulation},
  author={Maddukuri, Abhiram and Jiang, Zhenyu and Chen, Lawrence Yunliang and Nasiriany, Soroush and Xie, Yuqi and Fang, Yu and Huang, Wenqi and Wang, Zu and Xu, Zhenjia and Chernyadev, Nikita and others},
  journal={arXiv preprint arXiv:2503.24361},
  year={2025}
}

@inproceedings{long2015learning,
  title={Learning transferable features with deep adaptation networks},
  author={Long, Mingsheng and Cao, Yue and Wang, Jianmin and Jordan, Michael},
  booktitle={International conference on machine learning},
  pages={97--105},
  year={2015},
  organization={PMLR}
}

@article{gold2018dynamic,
  title={Dynamic time warping and geometric edit distance: Breaking the quadratic barrier},
  author={Gold, Omer and Sharir, Micha},
  journal={ACM Transactions On Algorithms (TALG)},
  volume={14},
  number={4},
  pages={1--17},
  year={2018},
  publisher={ACM New York, NY, USA}
}

@article{zhu2020robosuite,
  title={robosuite: A modular simulation framework and benchmark for robot learning},
  author={Zhu, Yuke and Wong, Josiah and Mandlekar, Ajay and Mart{\'\i}n-Mart{\'\i}n, Roberto and Joshi, Abhishek and Nasiriany, Soroush and Zhu, Yifeng},
  journal={arXiv preprint arXiv:2009.12293},
  year={2020}
}

@article{van2008visualizing,
  title={Visualizing data using t-SNE.},
  author={Van der Maaten, Laurens and Hinton, Geoffrey},
  journal={Journal of machine learning research},
  volume={9},
  number={11},
  year={2008}
}

@inproceedings{james2019sim,
  title={Sim-to-real via sim-to-sim: Data-efficient robotic grasping via randomized-to-canonical adaptation networks},
  author={James, Stephen and Wohlhart, Paul and Kalakrishnan, Mrinal and Kalashnikov, Dmitry and Irpan, Alex and Ibarz, Julian and Levine, Sergey and Hadsell, Raia and Bousmalis, Konstantinos},
  booktitle={Proceedings of the IEEE/CVF conference on computer vision and pattern recognition},
  pages={12627--12637},
  year={2019}
}

@inproceedings{qi2017pointnet,
  title={Pointnet: Deep learning on point sets for 3d classification and segmentation},
  author={Qi, Charles R and Su, Hao and Mo, Kaichun and Guibas, Leonidas J},
  booktitle={Proceedings of the IEEE conference on computer vision and pattern recognition},
  pages={652--660},
  year={2017}
}

@inproceedings{chi2023diffusionpolicy,
	title={Diffusion Policy: Visuomotor Policy Learning via Action Diffusion},
	author={Chi, Cheng and Feng, Siyuan and Du, Yilun and Xu, Zhenjia and Cousineau, Eric and Burchfiel, Benjamin and Song, Shuran},
	booktitle={Proceedings of Robotics: Science and Systems (RSS)},
	year={2023}
}

@inproceedings{he2016deep,
  title={Deep residual learning for image recognition},
  author={He, Kaiming and Zhang, Xiangyu and Ren, Shaoqing and Sun, Jian},
  booktitle={Proceedings of the IEEE conference on computer vision and pattern recognition},
  pages={770--778},
  year={2016}
}

@inproceedings{courty2014domain,
  title={Domain adaptation with regularized optimal transport},
  author={Courty, Nicolas and Flamary, R{\'e}mi and Tuia, Devis},
  booktitle={Machine Learning and Knowledge Discovery in Databases: European Conference, ECML PKDD 2014, Nancy, France, September 15-19, 2014. Proceedings, Part I 14},
  pages={274--289},
  year={2014},
  organization={Springer}
}

@inproceedings{redko2017theoretical,
  title={Theoretical analysis of domain adaptation with optimal transport},
  author={Redko, Ievgen and Habrard, Amaury and Sebban, Marc},
  booktitle={Machine Learning and Knowledge Discovery in Databases: European Conference, ECML PKDD 2017, Skopje, Macedonia, September 18--22, 2017, Proceedings, Part II 10},
  pages={737--753},
  year={2017},
  organization={Springer}
}

@article{perrot2016mapping,
  title={Mapping estimation for discrete optimal transport},
  author={Perrot, Micha{\"e}l and Courty, Nicolas and Flamary, R{\'e}mi and Habrard, Amaury},
  journal={Advances in Neural Information Processing Systems},
  volume={29},
  year={2016}
}

@inproceedings{zhao2019learning,
  title={On learning invariant representations for domain adaptation},
  author={Zhao, Han and Des Combes, Remi Tachet and Zhang, Kun and Gordon, Geoffrey},
  booktitle={International conference on machine learning},
  pages={7523--7532},
  year={2019},
  organization={PMLR}
}

@article{hogan1985impedance,
  title={Impedance control: An approach to manipulation: Part II—Implementation},
  author={Hogan, Neville},
  year={1985}
}

@article{tsai1989new,
  title={A new technique for fully autonomous and efficient 3 d robotics hand/eye calibration},
  author={Tsai, Roger Y and Lenz, Reimar K and others},
  journal={IEEE Transactions on robotics and automation},
  volume={5},
  number={3},
  pages={345--358},
  year={1989}
}

@article{han2023quickfps,
  title={QuickFPS: Architecture and Algorithm Co-Design for Farthest Point Sampling in Large-Scale Point Clouds},
  author={Han, Meng and Wang, Liang and Xiao, Limin and Zhang, Hao and Zhang, Chenhao and Xu, Xiangrong and Zhu, Jianfeng},
  journal={IEEE Transactions on Computer-Aided Design of Integrated Circuits and Systems},
  year={2023},
  publisher={IEEE}
}

@inproceedings{robomimic2021,
  title={What Matters in Learning from Offline Human Demonstrations for Robot Manipulation},
  author={Mandlekar, Ajay and Xu, Danfei and Wong, Josiah and Nasiriany, Soroush and Wang, Chen and Kulkarni, Rohun and Fei-Fei, Li and Savarese, Silvio and Zhu, Yuke and Mart{\'\i}n-Mart{\'\i}n, Roberto},
  booktitle={5th Annual Conference on Robot Learning},
  year={2021}
}

@inproceedings{kim2020domain,
  title={Domain adaptive imitation learning},
  author={Kim, Kuno and Gu, Yihong and Song, Jiaming and Zhao, Shengjia and Ermon, Stefano},
  booktitle={International Conference on Machine Learning},
  pages={5286--5295},
  year={2020},
  organization={PMLR}
}

@article{nguyen2024dude,
  title={Dude: Dual distribution-aware context prompt learning for large vision-language model},
  author={Nguyen, Duy MH and Le, An T and Nguyen, Trung Q and Diep, Nghiem T and Nguyen, Tai and Duong-Tran, Duy and Peters, Jan and Shen, Li and Niepert, Mathias and Sonntag, Daniel},
  journal={arXiv preprint arXiv:2407.04489},
  year={2024}
}

@inproceedings{raychaudhuri2021cross,
  title={Cross-domain imitation from observations},
  author={Raychaudhuri, Dripta S and Paul, Sujoy and Vanbaar, Jeroen and Roy-Chowdhury, Amit K},
  booktitle={International conference on machine learning},
  pages={8902--8912},
  year={2021},
  organization={PMLR}
}

@inproceedings{kedia2025one,
  title={One-shot imitation under mismatched execution},
  author={Kedia, Kushal and Dan, Prithwish and Chao, Angela and Pace, Maximus A and Choudhury, Sanjiban},
  booktitle={2025 IEEE International Conference on Robotics and Automation (ICRA)},
  pages={15649--15656},
  year={2025},
  organization={IEEE}
}

@article{dan2025x,
  title={X-Sim: Cross-Embodiment Learning via Real-to-Sim-to-Real},
  author={Dan, Prithwish and Kedia, Kushal and Chao, Angela and Duan, Edward Weiyi and Pace, Maximus Adrian and Ma, Wei-Chiu and Choudhury, Sanjiban},
  journal={arXiv preprint arXiv:2505.07096},
  year={2025}
}

\newpage
\newpage
\section*{NeurIPS Paper Checklist}

\begin{enumerate}

\item {\bf Claims}
    \item[] Question: Do the main claims made in the abstract and introduction accurately reflect the paper's contributions and scope?
    \item[] Answer: \answerYes{} % Replace by \answerYes{}, \answerNo{}, or \answerNA{}.
    \item[] Justification: The paper clearly states the claims made and contributions in the introduction and abstract.

\item {\bf Limitations}
    \item[] Question: Does the paper discuss the limitations of the work performed by the authors?
    \item[] Answer: \answerYes{} % Replace by \answerYes{}, \answerNo{}, or \answerNA{}.
    \item[] Justification: The limitation of our method is discussed in the conclusion section.

\item {\bf Theory assumptions and proofs}
    \item[] Question: For each theoretical result, does the paper provide the full set of assumptions and a complete (and correct) proof?
    \item[] Answer: \answerNA{} % Replace by \answerYes{}, \answerNo{}, or \answerNA{}.
    \item[] Justification: 

    \item {\bf Experimental result reproducibility}
    \item[] Question: Does the paper fully disclose all the information needed to reproduce the main experimental results of the paper to the extent that it affects the main claims and/or conclusions of the paper (regardless of whether the code and data are provided or not)?
    \item[] Answer: \answerYes{} % Replace by \answerYes{}, \answerNo{}, or \answerNA{}.
    \item[] Justification: The complete implementation detail is included in the Appendix

\item {\bf Open access to data and code}
    \item[] Question: Does the paper provide open access to the data and code, with sufficient instructions to faithfully reproduce the main experimental results, as described in supplemental material?
    \item[] Answer: \answerYes{} % Replace by \answerYes{}, \answerNo{}, or \answerNA{}.
    \item[] Justification: The code and data are available on our project webpage.

\item {\bf Experimental setting/details}
    \item[] Question: Does the paper specify all the training and test details (e.g., data splits, hyperparameters, how they were chosen, type of optimizer, etc.) necessary to understand the results?
    \item[] Answer: \answerYes{} % Replace by \answerYes{}, \answerNo{}, or \answerNA{}.
    \item[] Justification: We include full detail in Appendix

\item {\bf Experiment statistical significance}
    \item[] Question: Does the paper report error bars suitably and correctly defined or other appropriate information about the statistical significance of the experiments?
    \item[] Answer: \answerNo{} % Replace by \answerYes{}, \answerNo{}, or \answerNA{}.
    \item[] Justification: The main result of our paper requires substantial effort to execute each real world evaluation trial. Therefore, producing error bars is impractical. We provide detailed description on the evaluation conditions of our experiment.

\item {\bf Experiments compute resources}
    \item[] Question: For each experiment, does the paper provide sufficient information on the computer resources (type of compute workers, memory, time of execution) needed to reproduce the experiments?
    \item[] Answer: \answerYes{} % Replace by \answerYes{}, \answerNo{}, or \answerNA{}.
    \item[] Justification: We provide full detail on computation resource required for our experiments.
    
\item {\bf Code of ethics}
    \item[] Question: Does the research conducted in the paper conform, in every respect, with the NeurIPS Code of Ethics \url{https://neurips.cc/public/EthicsGuidelines}?
    \item[] Answer: \answerYes{} % Replace by \answerYes{}, \answerNo{}, or \answerNA{}.
    \item[] Justification: Our experiment does not involve formal human study. Demonstration data is collected by authors with explicit consent.

\item {\bf Broader impacts}
    \item[] Question: Does the paper discuss both potential positive societal impacts and negative societal impacts of the work performed?
    \item[] Answer: \answerNA{} % Replace by \answerYes{}, \answerNo{}, or \answerNA{}.
    \item[] Justification: Our work is largely an algorithm for improving robot performance with simulation data. We do not expect immediate and direct societal impact of our work.
    
\item {\bf Safeguards}
    \item[] Question: Does the paper describe safeguards that have been put in place for responsible release of data or models that have a high risk for misuse (e.g., pretrained language models, image generators, or scraped datasets)?
    \item[] Answer: \answerNA{} % Replace by \answerYes{}, \answerNo{}, or \answerNA{}.
    \item[] Justification: 

\item {\bf Licenses for existing assets}
    \item[] Question: Are the creators or original owners of assets (e.g., code, data, models), used in the paper, properly credited and are the license and terms of use explicitly mentioned and properly respected?
    \item[] Answer: \answerYes{} % Replace by \answerYes{}, \answerNo{}, or \answerNA{}.
    \item[] Justification: Our work is built on existing open source codebases, which we provide adequate citation and credit.

\item {\bf New assets}
    \item[] Question: Are new assets introduced in the paper well documented and is the documentation provided alongside the assets?
    \item[] Answer: \answerNA{} % Replace by \answerYes{}, \answerNo{}, or \answerNA{}.
    \item[] Justification: The paper does not release new assets.

\item {\bf Crowdsourcing and research with human subjects}
    \item[] Question: For crowdsourcing experiments and research with human subjects, does the paper include the full text of instructions given to participants and screenshots, if applicable, as well as details about compensation (if any)? 
    \item[] Answer: \answerNA{} % Replace by \answerYes{}, \answerNo{}, or \answerNA{}.
    \item[] Justification: 

\item {\bf Institutional review board (IRB) approvals or equivalent for research with human subjects}
    \item[] Question: Does the paper describe potential risks incurred by study participants, whether such risks were disclosed to the subjects, and whether Institutional Review Board (IRB) approvals (or an equivalent approval/review based on the requirements of your country or institution) were obtained?
    \item[] Answer: \answerNA{} % Replace by \answerYes{}, \answerNo{}, or \answerNA{}.
    \item[] Justification: 

\item {\bf Declaration of LLM usage}
    \item[] Question: Does the paper describe the usage of LLMs if it is an important, original, or non-standard component of the core methods in this research? Note that if the LLM is used only for writing, editing, or formatting purposes and does not impact the core methodology, scientific rigorousness, or originality of the research, declaration is not required.
    %this research? 
    \item[] Answer: \answerNA{} % Replace by \answerYes{}, \answerNo{}, or \answerNA{}.
    \item[] Justification: 

\end{enumerate}

% appendix
\clearpage
\newpage
\appendix
\section{Table of Contents}

The supplementary material has the following contents:

\begin{itemize}
    \item \textbf{Task and Hardware Setups} (Sec.~\ref{sec:task_setups}): Detailed descriptions of the hardware setups, dataset, and task settings.
    \item \textbf{Model and Training Details} (Sec.~\ref{sec:model_details}): Descriptions of the neural network architectures used in our experiments and the corresponding training procedures.
    \item \textbf{Ablation Study on Sampling Strategy} (Sec.~\ref{sec:ablation_samp}): Evaluation and analysis of different sampling strategies.
    \item \textbf{Additional Results} (Sec.~\ref{sec:more_res}): In-distribution evaluation results for both image-based and point cloud-based policies in the real world.
    \item \textbf{Transport Plan Visualization} (Sec.~\ref{sec:plan_viz}): Visualizations of the optimal transport plan for randomly sampled training batches.
    \item \textbf{Visualization of Latent Space} (Sec.~\ref{sec:latent_viz}): Visual comparisons of the learned latent spaces between our method and the \texttt{Co-training} baseline across additional tasks.
\end{itemize}

More video results and analysis can be found on our website: \url{https://ot-sim2real.github.io/}

\section{Task and Hardware Setups} \label{sec:task_setups}
To evaluate the effectiveness of our approach, we conduct comprehensive experiments on a suite of robotic tabletop manipulation tasks, covering both sim-to-sim and sim-to-real transfer scenarios. These tasks are designed to test the system’s ability to handle key challenges in robotic manipulation, including dense object interactions, long-horizon reasoning, and high-precision control:

\begin{itemize}
\item \texttt{Lift}: Grasp the rim of a mug and lift it vertically;
\item \texttt{BoxInBin}: Grasp a tall box and place it into a bin;
\item \texttt{Stack}: Grasp a small cube and stack it on top of a longer cuboid;
\item \texttt{Square}: Grasp the handle of a square-shaped object and insert it onto a peg;
\item \texttt{MugHang}: Grasp the rim of a mug and hang it on a mug tree using the handle;
\item \texttt{Drawer}: Open a drawer, grasp a coffee pod from the table, place it into the drawer, and close the drawer.
\end{itemize}

\subsection{Hardware Setups}

\begin{figure}[ht!]
    \centering
    \includegraphics[width=0.7\linewidth]{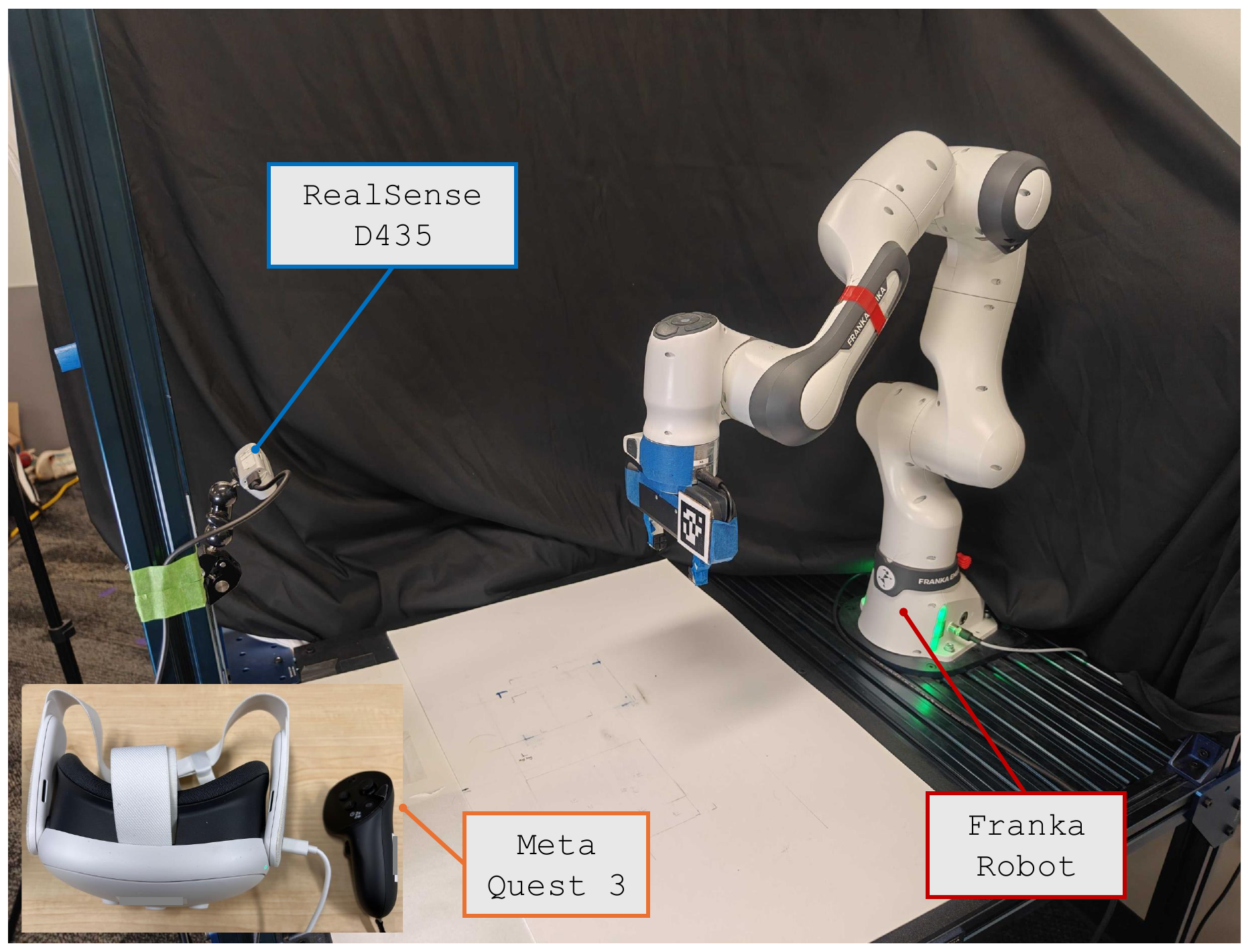}
    \caption{\textbf{Hardware Setup.} Our hardware platform uses a Franka Emika Panda robot, with an Intel RealSense D435 camera for capturing image and depth, and a Meta Quest 3 headset for teleoperation.}
    \label{fig:setup}
\end{figure}

 The system setup is illustrated in Fig.~\ref{fig:setup}. We use a Franka Emika Panda robot controlled via a joint impedance controller~\cite{hogan1985impedance} running at 20~Hz for policy execution. For data collection, the robot is teleoperated using a Meta Quest 3 headset, with tracked Cartesian poses converted to joint configurations through inverse kinematics. RGB image and depth are captured using an Intel RealSense D435 depth camera.

% \liqian{mention that we crop the scene point cloud using a bounding box}

\subsection{Domain Shifts and Observation Gaps}
We assess generalization under visual domain shifts in simulation through designing the following target domain shifts:

\begin{itemize}
    \item {\texttt{Viewpoint1-Point}}: The camera is rotated approximately $30^\circ$ around the z-axis, resulting in a side view in the target domain compared to a front-facing view in the source. Point cloud observations are used.
    
    \item {\texttt{Viewpoint3-Point}}: The camera is rotated approximately $90^\circ$ around the z-axis, introducing a more extreme viewpoint shift. Point cloud observations are used.
    
    \item {\texttt{Perturbation-Point}}: Random noise sampled uniformly from the range $[-0.01, 0.01]$ is added to each point in the point cloud to simulate sensor noise or domain shift.
    
    \item {\texttt{Viewpoint1-Image}}: A $20^\circ$ camera rotation around the z-axis is applied. RGB image observations are used.
    
    \item {\texttt{Texture-Image}}: The table texture in the target domain is modified. RGB image observations are used.
\end{itemize}

\begin{figure}
    \centering
    \includegraphics[width=0.8\linewidth]{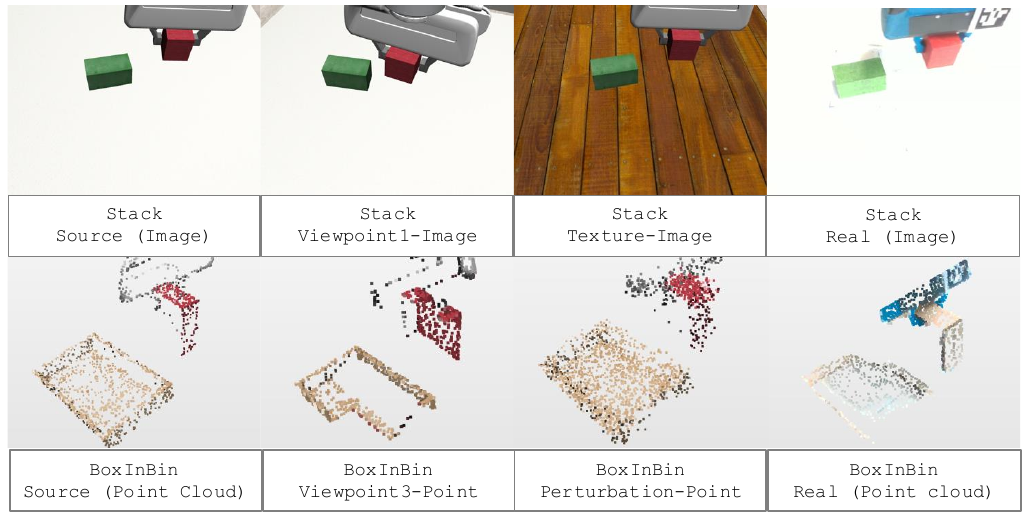}
    \caption{\textbf{Observation Gap Across Domains.} 
    % \danfei{This is very confusing --- top-bottom seems like pixel and point cloud of the same task... Can we label them in the images?} 
    Top: image observations for the \texttt{Stack} task from source, \texttt{Viewpoint1-Image}, \texttt{Texture-Image}, and real-world domains. Bottom: point cloud observations for the \texttt{BoxInBin} task from source, \texttt{Viewpoint3-Point}, \texttt{Perturbation-Point}, and real-world domains. Point cloud color is for visualization only and not used as input to the policy.}
    \label{fig:obs_gap}
    \vspace{-10pt}
\end{figure}

We illustrate the observation gap across all domains in Fig.~\ref{fig:obs_gap}. The first row displays image observations for the \texttt{Stack} task from the source domain, \texttt{Viewpoint1-Image}, \texttt{Texture-Image}, and the real world. The second row shows point cloud observations for the \texttt{BoxInBin} task from the source domain, \texttt{Viewpoint3-Point}, \texttt{Perturbation-Point}, and the real world. Point cloud color is for visualization only and not used as input to the policy.

\subsection{Task Datasets, Reset Ranges, and OOD Variants}\label{ssec:dataset_info}

% \shuo{list demo numbers for each task, show reset ranges for source and target demos}

% We focus primarily on evaluating policy performance in the region exclusively covered by the source demonstrations. To conduct controlled experiments, we define three distinct regions for each task: \texttt{Source}, \texttt{Target} In-Distribution (ID), and \texttt{Target-OOD}, as illustrated in Figure~\ref{fig:reset_range}. More specifically,
% \begin{itemize}
%     \item In the source domain, we generate 1000 demonstrations using MimicGen within the \texttt{Source} region.
%     \item In the target domain, we collect 10 demonstrations from the \texttt{Target} In-Distribution (ID) region for sim-to-sim transfer experiments. For sim-to-real transfer experiments, the number of demonstrations from the \texttt{Target} In-Distribution (ID) region is adjusted based on the difficulty of each task, as detailed in Table~\ref{tab:real_reset_range}. The \texttt{Target} In-Distribution (ID) region is sparsely covered by demonstrations and is therefore considered in-distribution (ID).
%     \item  No demonstrations are collected from the \texttt{Target-OOD} region, which is therefore considered out-of-distribution (OOD).
% \end{itemize}

\begin{figure}
    \centering
    \includegraphics[width=1\linewidth]{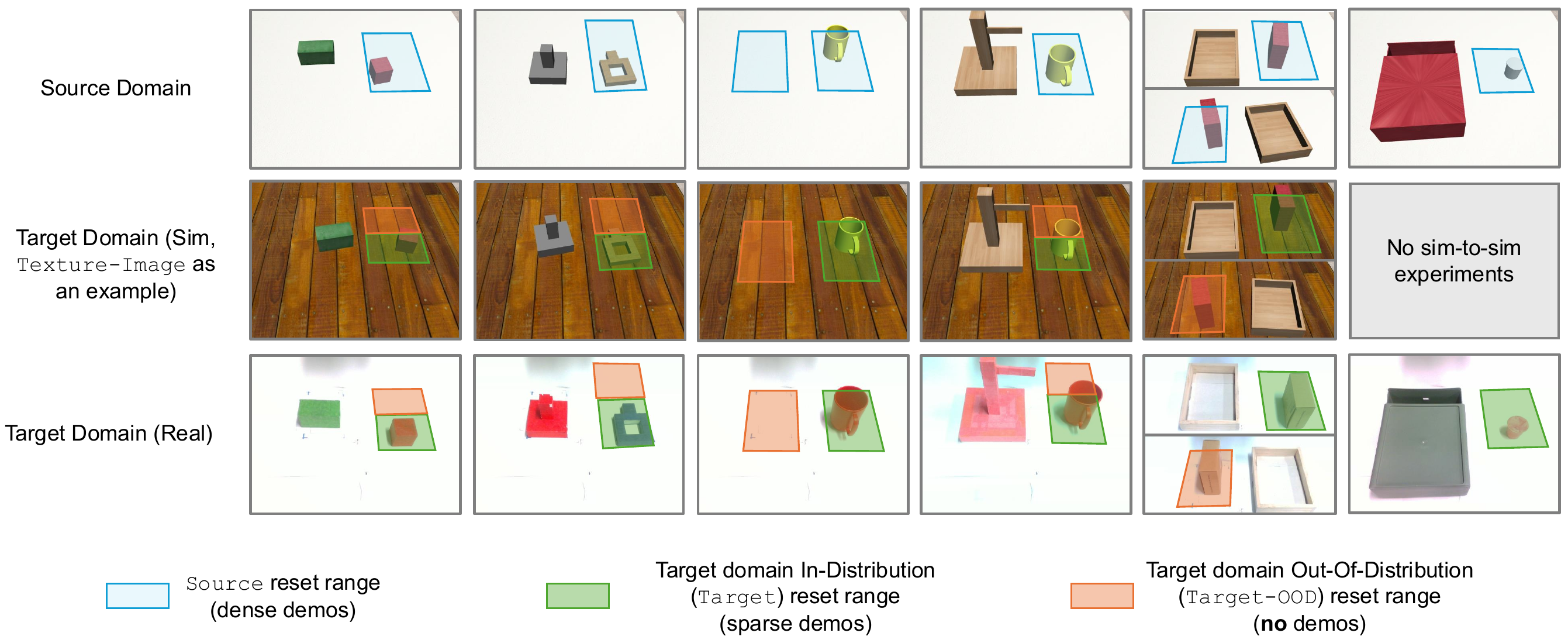}
    \caption{\textbf{Reset Ranges for Each Task.} The first row illustrates the \texttt{Source} region, where dense source-domain demonstrations are collected. The second row shows the \texttt{Target} and \texttt{Target-OOD} reset ranges used in sim-to-sim transfer experiments. In this setting, the \texttt{Target} region is sparsely covered by demonstrations, while the \texttt{Target-OOD} region contains no demonstrations and is used exclusively for policy evaluation. The third row similarly presents the \texttt{Target} and \texttt{Target-OOD} regions for sim-to-real transfer experiments.}
    \label{fig:reset_range}
\end{figure}

We focus primarily on evaluating policy performance in regions covered exclusively by source-domain demonstrations. To conduct controlled experiments, we define three distinct reset regions for each task—\texttt{Source}, \texttt{Target}, and \texttt{Target-OOD}—as shown in Fig.~\ref{fig:reset_range}. Specifically:

\begin{itemize}
\item \texttt{Source}: A large region that is densely covered by demonstrations in the source domain. We generate 1000 demonstrations using MimicGen~\cite{mandlekar2023mimicgen} within the \texttt{Source} region.

\item \texttt{Target}: A small subset of the \texttt{Source} region. This region is sparsely covered by demonstrations in the target domain, and is therefore considered in-distribution during evaluation. For sim-to-sim transfer, we collect 10 demonstrations within this region. For sim-to-real transfer, the number of real-world demonstrations collected in the \texttt{Target} region is adjusted based on task difficulty, as detailed in Tab.~\ref{tab:real_reset_range}.

\item \texttt{Target-OOD}: No demonstrations are collected in the \texttt{Target-OOD} region, which is used solely for evaluation and treated as out-of-distribution (OOD).
\end{itemize}

\begin{table}[]
\centering
\begin{tabular}{ccccccc}
\specialrule{1.pt}{0pt}{0pt}   % thick line before header
                & Lift & Stack & BoxInBin & MugHang & Square & Drawer \\ \hline
Number of real demos & 10 & 25 &  20 &  15 &  25    &  25  \\ \specialrule{1.pt}{0pt}{0pt}   % thick line before header
\end{tabular}
\caption{\textbf{Number of Real-World Demonstrations.} We collect 10–25 demonstrations per task, varying with task difficulty.}
\label{tab:real_reset_range}
\end{table}

\begin{figure}
    \centering
    \includegraphics[width=0.6\linewidth]{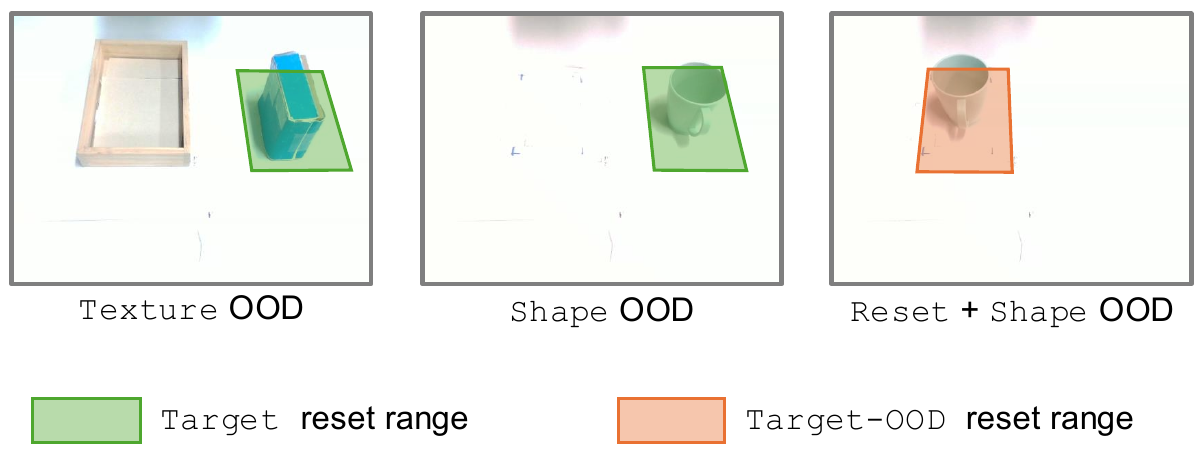}
    \caption{\textbf{Texture and Shape OOD in Sim-to-Real Experiments.} Visualization of reset ranges for the \texttt{BoxInBin} task under \texttt{Texture} OOD, and the \texttt{Lift} task under both \texttt{Shape} OOD and \texttt{Shape+Reset} OOD conditions.}\label{fig:shape_and_texture}
    \vspace{-20pt}
\end{figure}

For sim-to-real transfer experiments, in addition to the reset range OOD (denoted as \texttt{Reset}), we consider two additional OOD variants. In the \texttt{Texture} variant, the object's texture is modified to one that is unseen in the real-world demonstrations. In the \texttt{Shape} variant, the object is replaced with a novel shape not encountered in the real-world demonstrations. These variants are illustrated in Fig.~\ref{fig:shape_and_texture}.

% \shuo{show OOD variants \texttt{Texture}, \texttt{Shape}, \texttt{Reset}}

% \begin{figure}
%     \centering
%     \includegraphics[width=1\linewidth]{images/real_id.pdf}
%     \caption{}
%     \label{fig:real_id_setup}
% \end{figure}

% \begin{figure}
%     \centering
%     \includegraphics[width=1\linewidth]{images/real_ood.pdf}
%     \caption{Caption}
%     \label{fig:real_ood_setup}
% \end{figure}

% \subsection{Observation Gap Visualization}

% \liqian{merge with Simulation Domain Shifts?}

% \begin{figure}
%     \centering
%     \includegraphics[width=1\linewidth]{images/obs_gap.pdf}
%     \caption{\textbf{Observation Gap Across Domains.} Top: image observations for the \texttt{Stack} task from source, \texttt{Viewpoint1-Image}, \texttt{Texture-Image}, and real-world domains. Bottom: point cloud observations for the \texttt{BoxInBin} task from source, \texttt{Viewpoint3-Point}, \texttt{Perturbation-Point}, and real-world domains. Point cloud color is for visualization only and not used as input to the policy.}
%     \label{fig:obs_gap}
% \end{figure}

% We illustrate the observation gap across domains in Fig.~\ref{fig:obs_gap}. The first row displays image observations for the \texttt{Stack} task from the source domain, \texttt{Viewpoint1-Image}, \texttt{Texture-Image}, and the real world. The second row shows point cloud observations for the \texttt{BoxInBin} task from the source domain, \texttt{Viewpoint3-Point}, \texttt{Perturbation-Point}, and the real world. Point cloud color is for visualization only and not used as input to the policy.

\section{Model and Training Details}\label{sec:model_details}

For point cloud-based experiments, we adopt the 3D Diffusion Policy architecture~\cite{Ze2024DP3} with a PointNet encoder~\cite{qi2017pointnet}. The diffusion head receives features extracted from the point cloud observations along with robot proprioceptive inputs (joint and gripper positions), and outputs 7-DOF target joint positions and the gripper action. We project the depth map into the robot base frame to generate the scene point cloud. For a pixel with coordinate $(u, v)$ and depth $d$, the corresponding 3D location can be recovered by:

\[
p^w = R\cdot K^{-1}\cdot I + t
\]

where $I=(u\cdot d, v\cdot d, d)$, $[R\mid t]$ denotes the camera pose obtained through hand-eye calibration~\cite{tsai1989new}, and $K$ denotes the camera intrinsic matrix. We crop the reconstructed scene point cloud using a bounding box defined by \(x \in [-0.2, 0.1]\), \(y \in [-0.2, 0.2]\), and \(z\in [0.008, 0.588]\) to exclude irrelevant background information. The cropped point cloud is then downsampled to 2048 points using Farthest Point Sampling (FPS)~\cite{han2023quickfps}.

% \shuo{Model arch and training hyperparams}

% \subsection{Pixel-based experiments}
For experiments with image-based policy, we adopt Diffusion Policy~\cite{chi2023diffusion} with a ResNet-based~\cite{he2016deep} visual encoder. The original images are captured by the camera at a resolution of \(480 \times 640\). During preprocessing, the images are downsampled to \(120 \times 160\), followed by random cropping to \(108 \times 144\) during training and center cropping during testing. The policy takes stacked history images and robot proprioceptive inputs (joint and gripper positions) as input, and outputs 7-DOF target joint positions along with the gripper action.

Our overall training procedure is summarized in Algm.~\ref{algm:learning}. We use a batch size of 256 for the behavior cloning loss \(L_{\text{BC}}\), with a co-training ratio of 0.9 following Maddukuri et al.~\cite{maddukuri2025sim}. For the optimal transport loss \(L_{\text{OT}}\), the batch size is set to 128, with a weighting coefficient \(\lambda = 0.1\). We use \(\epsilon = 0.0005\) and \(\tau = 0.01\) in our experiments.

% Following~\cite{chi2023diffusion}, we set the prediction horizon, action horizon, and observation horizon to be 16, 8, and 2, respectively. 

%\begin{wrapfigure}{r}{0.5\textwidth}
%\vspace{-10pt} % adjust vertical spacing if needed
%\begin{minipage}{0.5\textwidth}
\begin{algorithm}[H]
\caption{Joint Policy Training with OT\label{algm:learning}}
\begin{algorithmic}[1]
    \Require Source dataset ${D}_{src}$, Target dataset ${D}_{tgt}$
    \State Initialize encoder $f_\phi$, and policy $\pi_\theta$
    \State Compute DTW distances for all trajectories pairs in ${D}_{src}$ and ${D}_{tgt}$
    
    \For{iteration $t = 1$ to $T$}
        \State Sample a paired batch $\{(o^{i}_{src}, x_{src}^{i}, a^{i}_{src}, o^{j}_{tgt}, x_{tgt}^{j}, a^{j}_{tgt})\}$ with size $N$ from ${D}_{src}$ and ${D}_{tgt}$ using strategy described in Sec.~\ref{ssec:sampling}
        \State Compute features $\{z^{i}_{src}\}$ and $\{z^{j}_{tgt}\}$ using encoder $f_\phi$
        \State Construct ground cost matrix $\hat{C}_{\phi}$ as described in Sec.~\ref{ssec:jot}
        \State Compute optimal transport plan ${\Pi^*} =\arg\min_{\Pi \in \mathbb{R}_{+}^{N\times N}} (\langle \Pi, \hat{C}_\phi \rangle_F + \epsilon \cdot \Omega(\Pi) + \tau \cdot \text{KL}(\Pi \mathbf{1} || \mathbf{p}) + \tau \cdot \text{KL}(\Pi^\top \mathbf{1} || \mathbf{q}) )$ via Sinkhorn-Knopp algorithm~\cite{cuturi2013sinkhorn} 
        \State Compute OT loss ${L}_{\text{UOT}}(f_{\phi}) = \langle{\Pi^*}, \hat{C}_{\phi}\rangle_{F}$ 
        \State Sample $\{(o^{i}_{src}, x_{src}^{i}, a^{i}_{src})\}$ from ${D}_{src}$ and sample $\{(o^{j}_{tgt}, x_{tgt}^{j}, a^{j}_{tgt})\}$ from ${D}_{tgt}$
        \State Compute BC loss ${L}_{\text{BC}}(f_{\phi}, \pi_{\theta})$ 
        \State Update $f_{\phi}$ and $\pi_\theta$ with gradients of ${L}_{\text{BC}}(f_{\phi}, \pi_{\theta})+\lambda\cdot{L}_\text{UOT}(f_{\phi})$
        % \State Update $f_\phi$ with gradients of $\mathcal{L}_{src}^{\text{BC}}+\mathcal{L}_{tgt}^{\text{BC}}+\lambda\cdot\mathcal{L}^{\text{OT}}$
        % \State \caelan{Do you do this update simultaneously or in separate (coordinate) steps?} \shuo{simultaneously}
    \EndFor

\end{algorithmic}
\end{algorithm}
%\end{minipage}
%\vspace{-10pt} % optional
%\end{wrapfigure}

\section{Ablation Study on Sampling Strategy}\label{sec:ablation_samp}

\begin{figure}[h]
    \centering
    \includegraphics[width=0.6\linewidth]{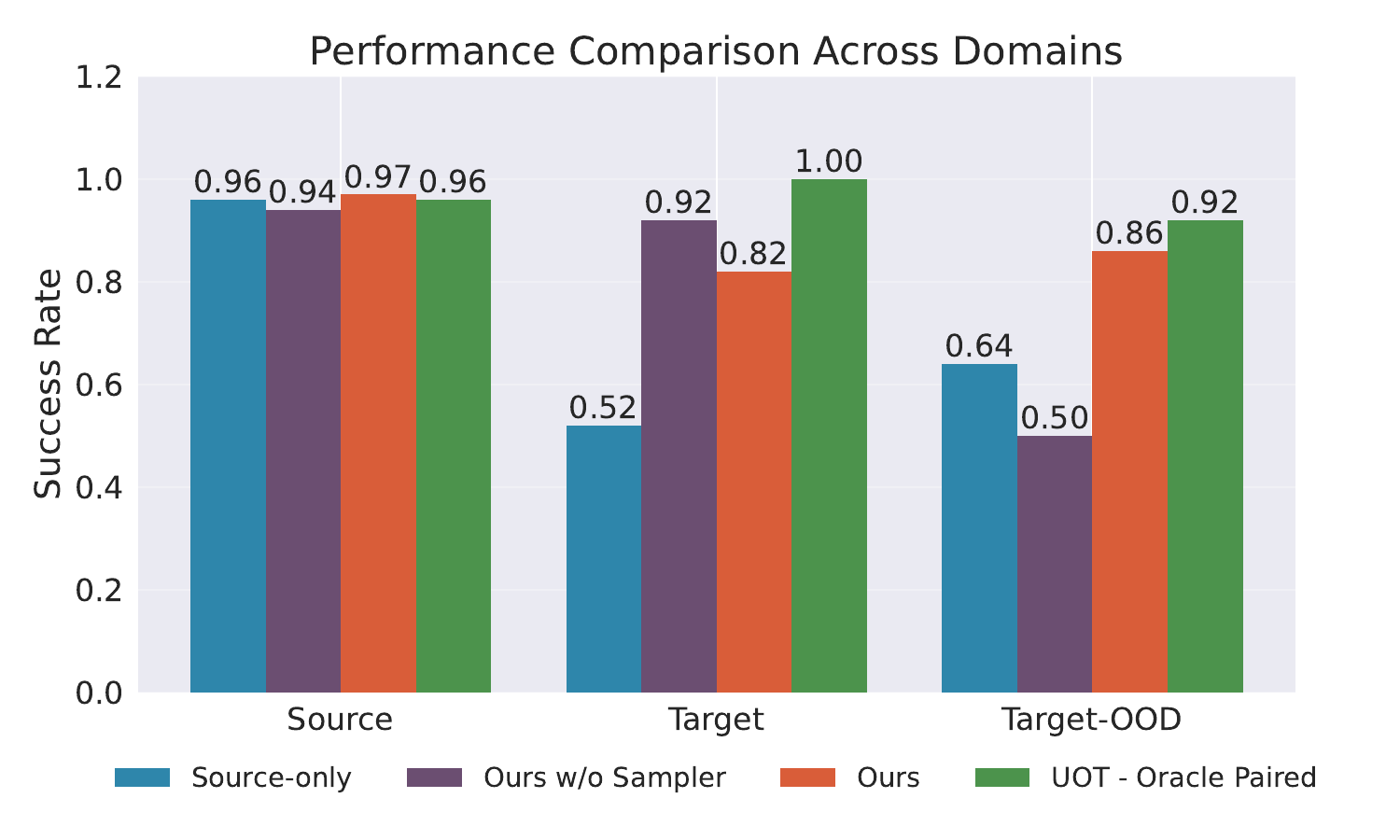}
    \caption{\textbf{Sampling Strategies Comparison.} Our proposed sampling strategy (\texttt{Ours}) improves policy success rates on the \texttt{Stack} task with \texttt{Viewpoint1-Point}, outperforming \texttt{Ours w/o Sampler}, and achieving performance comparable to the oracle-paired upper bound (\texttt{UOT-Oracle}).
    %We compare success rates of policies trained with different sampling strategies under unbalanced OT loss.
    % \caelan{Would move to the appendix to save space (not a clean win anyways)}
    % \shuo{May tune the colormap}
    }
    \label{fig:oracle_pairs}
\end{figure}
To assess the effectiveness of our sampling strategy, we compare the full method against a variant (denoted as \texttt{Ours w/o Sampler}) that does not apply any trajectory-level sampling. In this baseline, source and target data are randomly sampled across trajectories and time steps, with no coordination. We also include an oracle variant (denoted as \texttt{UOT-Oracle}), which constructs perfectly paired batches—each state is observed in both the source and target domains to ensure that batch data originates from the same underlying states. We evaluate policy performance on the \texttt{Stack} task under the \texttt{Viewpoint1-Point} variation, with results shown in Fig.~\ref{fig:oracle_pairs}.

\textbf{Temporal-aware strategy improves pairing quality and downstream performance.} The oracle baseline demonstrates that, given perfectly aligned data, unbalanced OT loss significantly enhances generalization by enabling the encoder to learn domain-invariant representations. In contrast, the no-sampling variant (\texttt{Ours w/o Sampler}) exhibits poor generalization in the \texttt{Target-OOD} setting. This degradation likely stems from the low probability of encountering aligned state pairs in mini-batches—especially problematic in long-horizon tasks, where uncoordinated sampling rarely produces temporally aligned data.

\section{Hyperparameter Sensitivity Analysis}
We conduct an ablation study to evaluate the sensitivity of our method to key hyperparameters, including the entropy regularization coefficient ($\epsilon$), the KL divergence penalty term ($\tau$), and the window size used in temporally aligned sampling. In each experiment, we vary a single hyperparameter while keeping the others fixed, train the policy, and assess its performance via rollouts. We report results for the \texttt{BoxInBin} task under the \texttt{Viewpoint-Image} setting and the \texttt{Lift} task under the \texttt{Texture-Image} setting, as shown in Table~\ref{tab:sensitivity}. The results indicate that our method is robust to hyperparameter variations within reasonable ranges. Specifically, performance remains stable when $\epsilon$ and $\tau$ are set between 0.001 and 0.1, and when the window size is varied between 5 and 20. Our method consistently outperforms the co-training baseline in OOD scenarios, where the baseline achieves a success rate of 0.14 on \texttt{BoxInBin} and 0.6 on \texttt{Lift}. These findings suggest that, although our approach introduces additional components, it does not require extensive tuning and offers clear advantages in terms of generalization.

\begin{table}[t]
\centering
\begin{subtable}[t]{.49\linewidth}
\centering
\adjustbox{max width=\linewidth}{
\begin{tabular}{lcccccc}
\toprule
\multicolumn{7}{c}{$\epsilon$} \\
\midrule
 & 0.0001 & 0.001 & 0.005 & 0.01 & 0.04 & 1 \\
T   & 0.90 & 0.94 & 0.92 & 0.88 & 0.90 & 0.88 \\
T-O & 0.18 & 0.16 & 0.26 & 0.22 & 0.18 & 0.20 \\
\midrule
\multicolumn{7}{c}{$\tau$} \\
\midrule
 & 0.0001 & 0.001 & 0.005 & 0.02 & 0.04 & 1 \\
T   & 0.88 & 0.96 & 0.94 & 0.94 & 0.92 & 0.94 \\
T-O & 0.28 & 0.26 & 0.20 & 0.28 & 0.22 & 0.22 \\
\midrule
\multicolumn{7}{c}{winsize} \\
\midrule
 & 1 & 5 & 10 & 20 & 40 & 120 \\
T   & 0.82 & 0.92 & 0.86 & 0.90 & 0.94 & 0.84 \\
T-O & 0.20 & 0.22 & 0.22 & 0.24 & 0.16 & 0.14 \\
\bottomrule
\end{tabular}}
\caption{\texttt{BoxInBin}}
\end{subtable}\hfill
\begin{subtable}[t]{.49\linewidth}
\centering
\adjustbox{max width=\linewidth}{
\begin{tabular}{lcccccc}
\toprule
\multicolumn{7}{c}{$\epsilon$} \\
\midrule
 & 0.0001 & 0.001 & 0.01 & 0.04 & 0.1 & 1 \\
T   & 0.84 & 0.88 & 0.80 & 0.76 & 0.78 & 0.76 \\
T-O & 0.60 & 0.74 & 0.62 & 0.66 & 0.68 & 0.54 \\
\midrule
\multicolumn{7}{c}{$\tau$} \\
\midrule
 & 0.0001 & 0.005 & 0.02 & 0.04 & 0.1 & 1 \\
T   & 0.78 & 0.70 & 0.76 & 0.76 & 0.78 & 0.74 \\
T-O & 0.56 & 0.67 & 0.64 & 0.66 & 0.62 & 0.66 \\
\midrule
\multicolumn{7}{c}{winsize} \\
\midrule
 & 1 & 5 & 10 & 20 & 40 & 120 \\
T   & 0.86 & 0.80 & 0.70 & 0.78 & 0.74 & 0.82 \\
T-O & 0.66 & 0.60 & 0.67 & 0.58 & 0.56 & 0.60 \\
\bottomrule
\end{tabular}}
\caption{\texttt{Lift}}
\end{subtable}
\caption{\textbf{Hyperparameter Sensitivity}. In each series of experiments, we vary a single hyperparameter while keeping the others fixed, train the policy, and assess its success rates via rollouts. T and T-O denote the target domain and the target domain under OOD conditions.}
\label{tab:sensitivity}
\end{table}

\section{Additional Real-world Evaluation Results}\label{sec:more_res}

We conduct extensive real-world evaluations to validate the effectiveness of our approach. Sim-to-real transfer results for in-distribution scenarios are reported in Tab.\ref{exp:img_real_id} and Tab.\ref{exp:pc_real_id} for image-based and point cloud-based policies, respectively. Results for out-of-distribution (OOD) scenarios are presented in the main paper.

\begin{table}[!htbp]
\centering
\begin{tabular}{l|cc|cc|cc|c}
\specialrule{1.pt}{0pt}{0pt}   % thick line before header
 & \multicolumn{2}{c|}{\shortstack{\texttt{Stack}}}
 & \multicolumn{2}{c|}{\shortstack{\texttt{Square}}}
 & \multicolumn{2}{c|}{\shortstack{\texttt{BoxInBin}}}
 & \textbf{Average} \\ 
 & grasp & full & grasp & full & grasp & full & full \\
\midrule
\texttt{Source-only}   & 0.1 & 0.0 & 0.0 & 0.0 & 0.0 & 0.0 & 0.00 \\
 \texttt{Target-only}  & 0.7 & 0.7 & 0.8 & 0.0 & 0.7 & 0.7 & 0.47 \\
\texttt{Co‑training}   & 0.8 & 0.7 & 0.8 & 0.1 & \textbf{0.9} & 0.8 & 0.53 \\
\texttt{Ours}       & \textbf{0.9} & \textbf{0.9} & \textbf{0.9} & \textbf{0.4} & \textbf{0.9} & \textbf{0.9} & \textbf{0.73} \\
\specialrule{1.pt}{0pt}{0pt}   % thick line after data
\end{tabular}
\caption{\textbf{Real World Image-Based Policy In-Distribution Success Rates.} The \textbf{Average} denotes the average full task success rates over all tasks.
% \caelan{Move to appendix}
% \caelan{Use texttt for \texttt{co-train} and other approaches for consistency}
}\label{exp:img_real_id}
\end{table}

\begin{table}[!htbp]
\centering
\setlength{\tabcolsep}{2.5pt}
\begin{tabular}{l|cc|cc|cc|cc|cc|cccc|c}
\specialrule{1.pt}{0pt}{0pt}   % thick line before header
 & \multicolumn{2}{c|}{\shortstack{\texttt{Stack}}} 
 & \multicolumn{2}{c|}{\shortstack{\texttt{Square}}} 
 & \multicolumn{2}{c|}{\shortstack{\texttt{BoxInBin}}} 
 & \multicolumn{2}{c|}{\shortstack{\texttt{Lift}}} 
 & \multicolumn{2}{c|}{\shortstack{\texttt{MugHang}}} 
 & \multicolumn{4}{c|}{\shortstack{\texttt{Drawer}}} 
 & \textbf{Average} \\
 & grasp & full & grasp & full & grasp & full & reach & full & grasp & full 
 & open & grasp & place & full 
 & full \\
\midrule
\texttt{S.-only} & 0.3 & 0.0 & 0.1 & 0.1 & 0.4 & 0.3 & 0.5 & 0.5 & 0.1 & 0.0 
           & 0.0 & 0.0 & 0.0 & 0.0 & 0.15 \\
\texttt{T.-only} & 0.7 & 0.4 & 0.6 & 0.1 & \textbf{0.9} & 0.8 & \textbf{1.0} & \textbf{1.0} 
           & 0.8 & \textbf{0.8} & 0.9 & 0.5 & 0.5 & 0.5 & 0.60 \\
\texttt{Co‑train.}   & 0.7 & 0.7 & \textbf{1.0} & \textbf{0.5} & 0.8 & 0.8 & 0.8 & 0.8 
           & \textbf{1.0} & \textbf{0.8} & \textbf{1.0} & \textbf{0.7} & \textbf{0.7} & 0.4 & 0.67 \\
\texttt{Ours}      & \textbf{0.8} & \textbf{0.8} & \textbf{1.0} & 0.4 & \textbf{0.9} & \textbf{0.9} 
           & \textbf{1.0} & \textbf{1.0} & \textbf{1.0} & \textbf{0.8} 
           & \textbf{1.0} & \textbf{0.7} & \textbf{0.7} & \textbf{0.7} & \textbf{0.77} \\
\specialrule{1.pt}{0pt}{0pt}   % thick line after data
\end{tabular}
\caption{\textbf{Real World Point-Cloud-Based Policy In-Distribution Success Rates.} The \textbf{Average} denotes the average full task success rates over all tasks. \label{exp:pc_real_id}
% \caelan{Move to appendix}
}
\end{table}

Experimental results show that our approach consistently outperforms all baselines in real-world in-distribution settings. Our method achieves average success rates of 0.73 and 0.77 for image-based and point cloud-based policies, respectively, demonstrating its effectiveness in learning complex real-world manipulation tasks.

\section{Performance with Scarce Target Domain Data}

% \begin{table}[h]
% \centering
% \resizebox{\columnwidth}{!}{%
% \begin{tabular}{|cc|c|c|c|c|}
% \hline
%                                               &     & Ours & Co-t. & MMD  & Tgt-only \\ \hline
% \multicolumn{1}{|c|}{\multirow{2}{*}{1 Demo}} & T   & 0.56 & 0.46  & 0.42 & 0.00     \\ \cline{2-6} 
% \multicolumn{1}{|c|}{}                        & T-O & 0.28 & 0.00  & 0.16 & 0.00     \\ \hline
% \multicolumn{1}{|c|}{\multirow{2}{*}{5 Demo}} & T   & 0.70 & 0.38  & 0.34 & 0.46     \\ \cline{2-6} 
% \multicolumn{1}{|c|}{}                        & T-O & 0.32 & 0.22  & 0.22 & 0.00     \\ \hline
% \end{tabular}%
% }
% \end{table}

\begin{table}[!htbp]
\centering
\begin{tabular}{l|cc|cc}
\specialrule{1.pt}{0pt}{0pt}   % thick line before header
 & \multicolumn{2}{c|}{\shortstack{1 Demo}} 
 & \multicolumn{2}{c}{\shortstack{5 Demo}} \\ 
 & Target & Target-OOD & Target & Target-OOD \\
\midrule
\texttt{Ours}       & \textbf{0.56} & \textbf{0.28} & \textbf{0.70} & \textbf{0.32} \\
\texttt{Co-training} & 0.46 & 0.00 & 0.38 & 0.22 \\
\texttt{MMD}        & 0.42 & 0.16 & 0.34 & 0.22 \\
\texttt{Target-only} & 0.00 & 0.00 & 0.46 & 0.00 \\
\specialrule{1.pt}{0pt}{0pt}   % thick line after data
\end{tabular}
\caption{\textbf{Performance with Limited Target Domain Data.} We report success rates for various methods on the \texttt{BoxInBin} task with the \texttt{Viewpoint3-Point} setting, under scenarios where data from the target domain is extremely limited.}
\label{tab:limited_data}
\end{table}

To assess the effectiveness of our method under limited target domain data, we compare it against several baselines in the low-data regime for the \texttt{BoxInBin} task with the \texttt{Viewpoint3-Point} setting. As shown in Tab.~\ref{tab:limited_data}, our approach consistently outperforms the baselines, highlighting its robustness even with extremely limited supervision.

\section{Transport Plan Visualization}\label{sec:plan_viz}

\begin{figure}[h]
    \centering
    \includegraphics[width=1.0\linewidth]{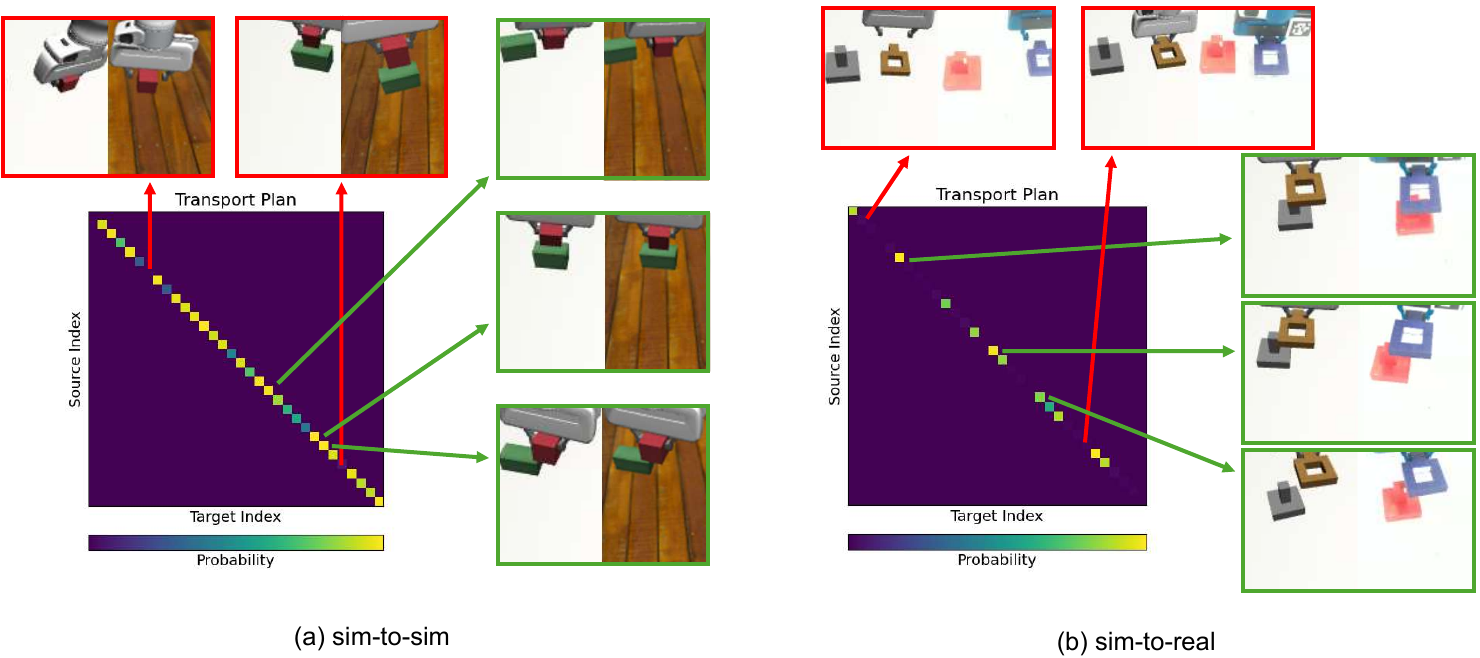}
    \caption{\textbf{Transport Plan Visualization.} We visualize the transport plan for a randomly sampled batch during training the image-based policy, alongside corresponding observations from both domains. The left figure shows a sim-to-sim experiment, while the right shows a sim-to-real experiment. The visualization reveals that the transport plan effectively aligns similar states across domains, as indicated by high transport probabilities.}
    % \caelan{Performance is too generic. Say success rate instead.}
    \label{fig:transport-plan}
\end{figure}

To understand how optimal transport facilitates domain-invariant feature learning and enhances cross-domain generalization, we visualize the transport plan for a randomly sampled batch during training the image-based policy, along with corresponding observations from both domains (see Fig.~\ref{fig:transport-plan}). The left plot shows results from the sim-to-sim transfer experiment, while the right plot depicts the sim-to-real setting. The results show that the transport plan effectively aligns similar states across domains, encouraging domain-invariant representations.

% \shuo{May also add a batch between sim and real?}

\section{Visualization of Latent Space}\label{sec:latent_viz}

\begin{figure}[h]
    \centering
    \includegraphics[width=1.0\linewidth]{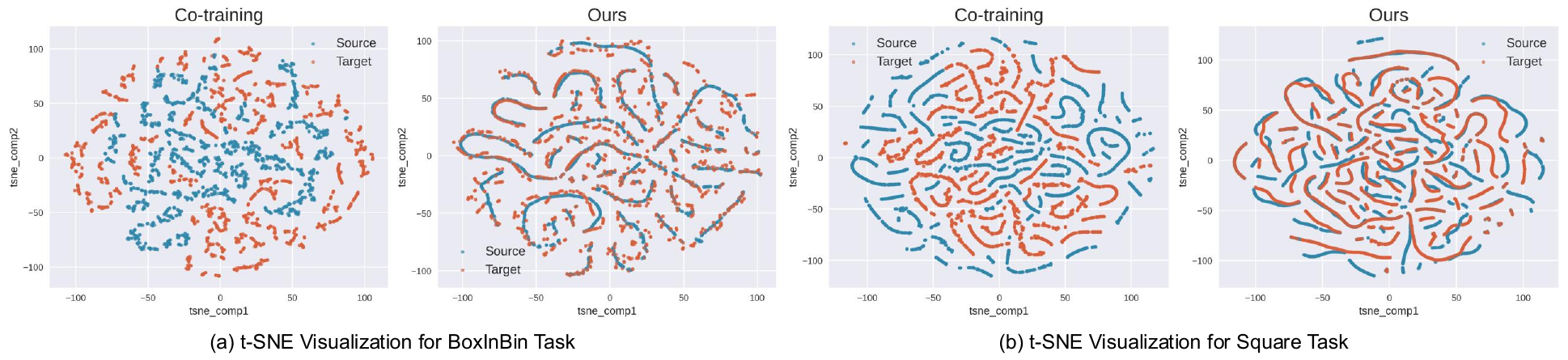}
    \caption{\textbf{Latent Space Visualization.} Latent space comparison between the \texttt{Co-training} baseline and our method. In our approach, source-domain points (blue) and target-domain points (red) form a well-mixed cluster, illustrating how OT alignment harmonizes cross-domain feature distributions and enhances transferability and generalization.}
    % \caelan{Performance is too generic. Say success rate instead.}
    \label{fig:more_tsne}
\end{figure}

Beyond the feature visualization for the \texttt{Stack} task with the \texttt{Viewpoint1-Point} target domain, we also present additional t-SNE~\cite{van2008visualizing} visualizations in Fig.~\ref{fig:more_tsne} for the \texttt{BoxInBin} task with the \texttt{Perturbation-Point} target domain, and the \texttt{Square} task with the \texttt{Viewpoint1-Point} target domain. We compare the latent spaces produced by the \texttt{Co-training} baseline and our method. In our approach, source-domain points (blue) and target-domain points (red) form a well-mixed cluster, highlighting how OT alignment harmonizes cross-domain feature distributions and improves both transferability and generalization.

% \newpage
% \input{text/rebuttal}

\end{document}